\documentclass[lettersize,journal]{IEEEtran}
\usepackage{amsmath,amsfonts}
\usepackage{algorithmic}
\usepackage{algorithm}
\usepackage{array}
\usepackage[caption=false,font=normalsize,labelfont=sf,textfont=sf]{subfig}
\usepackage{textcomp}
\usepackage{stfloats}
\usepackage{url}
\usepackage{verbatim}
\usepackage{graphicx}
\usepackage{cite}
\usepackage{amsmath}
\newcommand{\removelatexerror}{\let\@latex@error\@gobble}
\usepackage{threeparttable}
\begin{document}

\title{Siamese Transition Masked Autoencoders as Uniform Unsupervised Visual Anomaly Detector}

\author{Haiming Yao,~\IEEEmembership{Student Member,~IEEE,} Xue Wang,~\IEEEmembership{Senior Member,~IEEE, Wenyong Yu~\IEEEmembership{Member,~IEEE,}}

        % <-this % stops a space
\thanks{Manuscript received XX XX, 20XX; revised XX XX, 20XX. This study was supported in part by the XX XX (Grant No. XX XX), in part by the National Natural Science Foundation of China (Grant No. 51775214) (Corresponding author: Xue Wang.)
}% <-this % stops a space
\thanks{Haiming Yao and Xue Wang are with the State Key Laboratory of Precision Measurement Technology and Instruments, Department of Precision Instrument, Tsinghua University, Beijing 100084, China (e-mail:yhm22@mails.tsinghua.edu.cn, wangxue@mail.tsinghua.edu.cn).}
\thanks{Wenyong Yu is with with the School of Mechanical Science and Engineering, Huazhong University of Science and Technology, Wuhan 430074, China  (e-mail:ywy@hust.edu.cn).}
}

% The paper headers
\markboth{Journal of \LaTeX\ Class Files,~Vol.~14, No.~8, August~2021}%
{Shell \MakeLowercase{\textit{et al.}}: A Sample Article Using IEEEtran.cls for IEEE Journals}

\maketitle
{}
\begin{abstract}
Unsupervised visual anomaly detection conveys practical significance in many scenarios and is a challenging task due to the unbounded definition of anomalies. Moreover, most previous methods are application-specific, and establishing a unified model for anomalies across application scenarios remains unsolved. This paper proposes a novel hybrid framework termed Siamese Transition Masked Autoencoders(ST-MAE) to handle various visual anomaly detection tasks uniformly via deep feature transition. Concretely, the proposed method first extracts hierarchical semantics features from a pre-trained deep convolutional neural network and then develops a feature decoupling strategy to split the deep features into two disjoint feature patch subsets. Leveraging the decoupled features, the ST-MAE is developed with the Siamese encoders that operate on each subset of feature patches and perform the latent representations transition of two subsets, along with a lightweight decoder that reconstructs the original feature from the transitioned latent representation. Finally, the anomalous attributes can be detected using the semantic deep feature residual. Our deep feature transition scheme yields a nontrivial and semantic self-supervisory task to extract prototypical normal patterns, which allows for learning uniform models that generalize well for different visual anomaly detection tasks. The extensive experiments conducted demonstrate that the proposed ST-MAE method can advance state-of-the-art performance on multiple benchmarks across application scenarios with a superior inference efficiency, which exhibits great potential to be the uniform model for unsupervised visual anomaly detection.
\end{abstract}

\begin{IEEEkeywords}
Uniform visual anomaly detection, Anomaly localization, Siamese transition Masked Autoencoder
\end{IEEEkeywords}

\section{Introduction}

\IEEEPARstart{V}{isual} anomaly detection(VAD) is often referred to as identifying observations whose behavior deviates from a learned regular pattern. Due to the property of an automatic model to discriminate unknown samples, VAD has extensive applications in the real world, ranging from defect detection\cite{r1} in the industrial field to video surveillance in the smart city\cite{r30} to medical image analysis\cite{r31}, which exhibits significant importance in these settings. Owing to the uncertain nature of anomalies, obtaining and labeling enough novel examples is almost inaccessible. Accordingly, current works address this topic mainly following the unsupervised learning fashion in which the assessment of anomalies is modeled as a departure from intrinsically normal patterns.

\begin{figure}[t]\centering
\includegraphics[width=8.8cm]{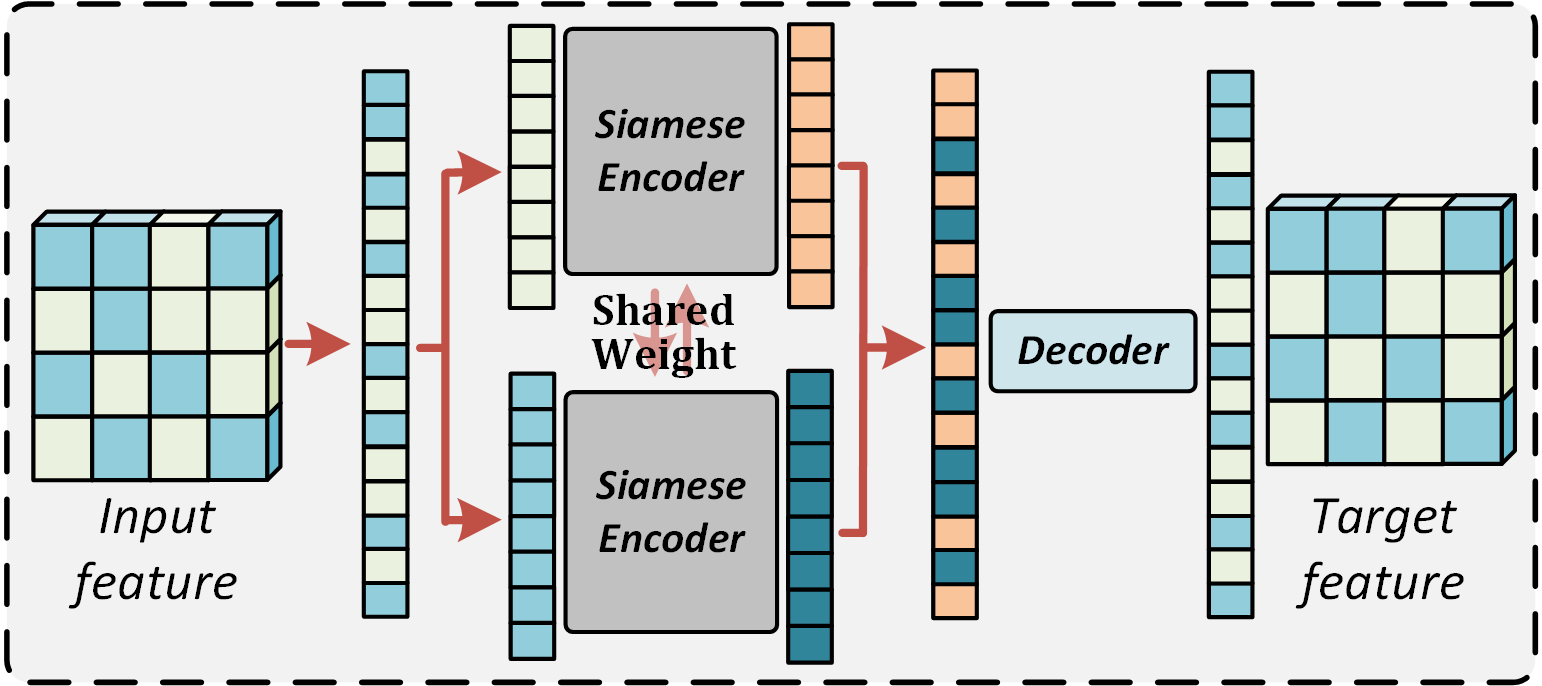}
\caption{Our ST-MAE architecture. The input feature is randomly masked and divided into two subsets of patches. The Siamese encoders are applied to each subset of visible patches. The encoded feature patch tokens of two subsets are transitioned in the sequence position after the encoder and reassembled as the full set of encoded feature patch tokens, which is finally processed by a small decoder to reconstruct the original feature.}
\label{FIG_1}
\end{figure}

In the context of VAD, unsupervised visual anomaly inspection and localization are two main tasks, and generally, the type of anomaly varies with the practical scenario. The classical anomalies can be divided into two categories: the overall shift in semantics or the localized regions suffering from structural or textural degradation. The former is usually conducted by the anomaly inspection approaches while the latter is generally solved by the anomaly localization methods. The visual anomaly inspection approaches\cite{r32} that typically construct a decision boundary between normal and anomalous data are commonly used to address the overall semantics anomaly. These techniques are widely utilized in the VAD community as the long-standing baselines which are, unfortunately, carry out in an image-level inference and frequently fall short of locating the abnormal region. While visual anomaly localization, the pixel-wise discriminant methods, are more helpful in real-world applications. A vast and expanding substantial literature has investigated this topic, mainly on embedding and reconstruction-based methods\cite{r34}. Embedding-based systems transform the raw image into an embedded high-dimensional feature space where the abnormality is highly discriminative from normal. Benefiting from representative feature extraction, embedding methods are superior in accuracy but typically come at a high cost in terms of online memory and inefficient inference. On the other hand, the reconstruction-based methods are more practical and have shown great promise for VAD. Deep autoencoder(AE)\cite{r33} is generally employed for this scheme, which only uses normal samples as training data and tries to reconstruct the input. In the testing phase, the reconstruction error is adopted as the criterion for anomaly identification. However, it has been observed that the vanilla AE can also reconstruct the anomalous well and produce a slight gap between the normal and anomalous samples, which is the bottleneck of detection performance. 

Many further extensions have been suggested to enhance the capabilities of the above methods but are still not superior enough in the detection efficiency and accuracy trade-off. In addition, despite the growing interest in VAD, the recent progress of reconstruction-based methods in VAD lags behind the embedding-based methods. We attempt to analyze this phenomenon from the major limitations of the following perspectives:(1) The pixel comparison from the raw image data lacks adequate high-level semantic and descriptive information, leading to an illegible difference\cite{r6}. (2) The slight discrepancy between the normal and abnormal samples, which can be attributed to the identity mapping of the trivial shortcut \cite{r8}, results in low discriminatory anomaly identification. More critically, most conventional methods are scenario-specific and only achieve acceptable performance on anomaly inspection or localization. To settle the above problems, in this paper, we propose the concept of uniform visual anomaly detection(UVAD), in which the vision-mediated anomaly inspection and localization can be simultaneously settled using one unified model.

\begin{figure}[t]\centering
\includegraphics[width=8.8cm]{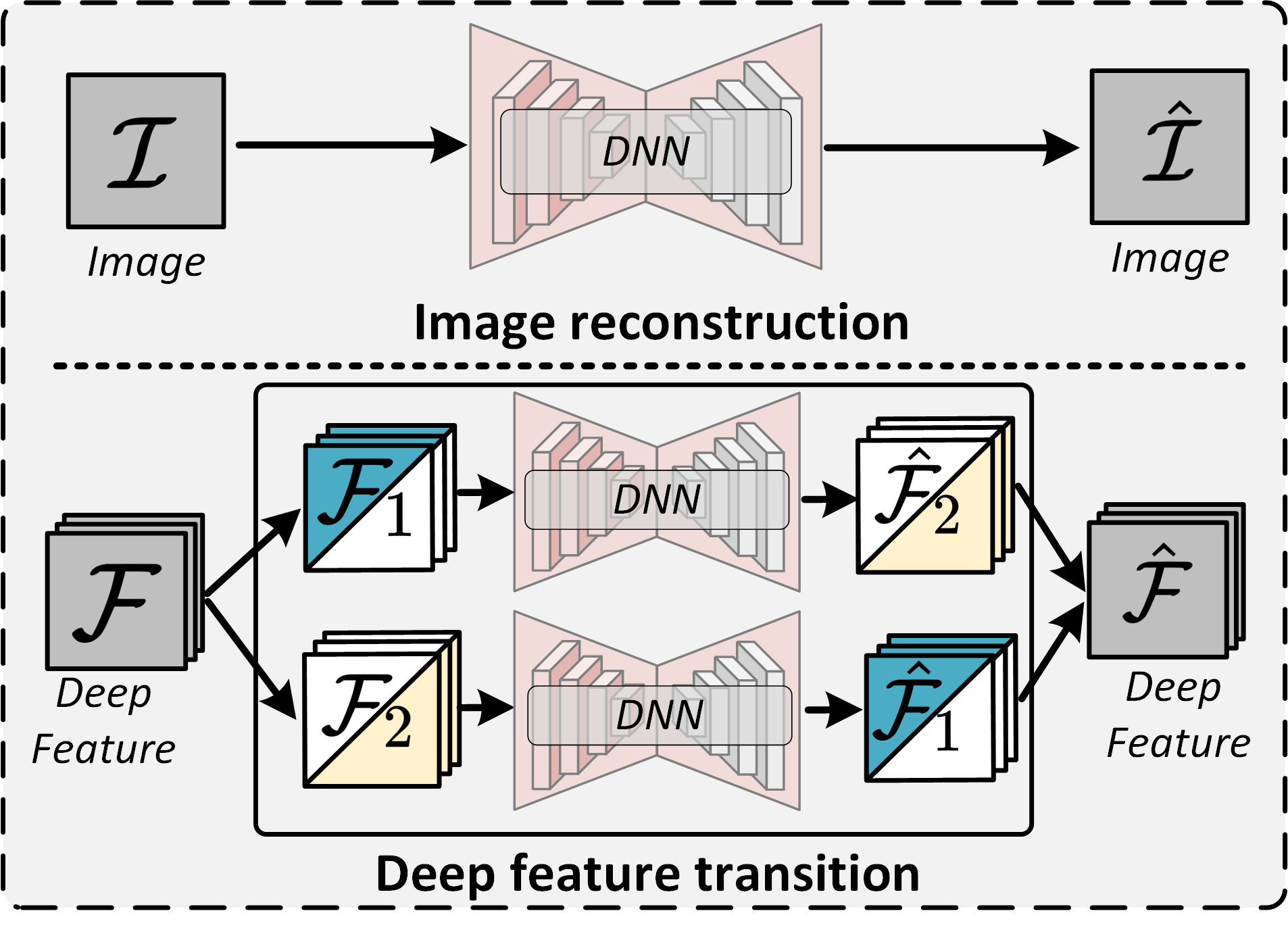}
\caption{Typical image reconstruction baseline versus the proposed deep feature transition. Autoencoders directly learn to reconstruct input images with the deep neural network(DNN) in pixels. While the proposed deep feature transition schema leverages the inspiration of Split-Brain\cite{r62} to convert the trivial pixel reconstruction to semantic deep feature transition.}
\label{FIG_1}
\end{figure}

We focus on reconstruction-based methods by proposing the heuristic paradigm: deep feature transition. As shown in Fig. 2, the typical pipeline of traditional reconstruction-based methods(Generally AE) and the main motivation of the proposed deep feature transition are illustrated. The vanilla AE framework is trained by reconstructing the normal image pixel and detecting anomalies based on the image reconstruction error. Instead of tackling VAD with straightforward image reconstruction, we further explore the reconstruction-based system by the deep discriminative feature transition, which learns representative semantics features under a self-supervised learning paradigm. Specifically, drawing on the advantages of embedding-based methods, we apply the deep features of the image with semantics context as the pattern descriptor rather than the low-level image pixel. Moreover, a split mechanism is added in the process, resulting in two disjoint subsets of deep features, and a Siamese architecture is adopted to transit one feature subset from another, thereby changing the problem from reconstruction to transition, which avoids the shortcut solution and facilitates capturing high-level semantics of normal patterns. Next, the two transitioned feature subsets are reassembled as the reconstructed results. Finally, the semantic feature residual instead of the pixel reconstruction error can be employed as the robust anomaly assessment.

Driven by the above analysis, we present a simple, yet effective, uniform approach termed Siamese transition masked autoencoder (ST-MAE) for UVAD. Concretely, our ST-MAE is proposed by leveraging the advantages of both embedding and reconstruction-based methods for the UVAD system. As depicted in Fig. 1, in analogy to MAE\cite{r61}, our ST-MAE has an asymmetric encoder-decoder design that reconstructs the input in feature space, yet our encoder applies a Siamese architecture. The input feature is randomly decoupled into two disjoint patch subsets, and each encoder operates only on the visible patches of each subset. Then, the full set of latent representations is reassembled by the encoded feature tokens of two subsets. Note that the relative positions of the feature tokens of the two subsets in the sequence of the full set are mutually transitioned during the reassembly phase. Finally, a lightweight decoder is employed to reconstruct the input feature from the full set of transitioned latent representations. Fig.3 shows the anomaly detection tasks for various application scenarios, namely the UVAD problem, and the main properties and qualitative results obtained by the proposed ST-MAE framework.

To summarize, our proposal contributes to the VAD community by designing a Siamese transition masked autoencoder(ST-MAE), the core contributions are two-fold: 
\begin{enumerate}
	\item We proposed the novel hybrid framework ST-MAE for UVAD by leveraging the deep feature transition, which can simultaneously tackle the problems of visual inspection and localization. Taking advantage of embedding-based and reconstruction-based approaches, ST-MAE can avoid the low-semantic and trivial shortcut solution in typical reconstruction-based systems and learn semantic prototypical representations for normal patterns. With the Siamese transition design, our ST-MAE has strong generalizability for various types of VAD scenarios.
 
	\item We experimented with ST-MAE in a series of comprehensive application scenarios, including the semantic one-class visual novelty detection task, the industrial scenarios defect detection tasks, the medical lesion detection task, and the video anomaly event detection. The results demonstrate that ST-MAE achieves state-of-the-art performance and reveals a great adaptation ability for extensive VAD tasks with superior inference efficiency. Moreover, we also extend our model to the few-shot learning for VAD. In this task, we observed that our ST-MAE can quickly adapt to novel categories with a considerable performance using a limited sample.
 
\end{enumerate}

The remainder of this article is organized as follows: Additional background and the works related to the VAD and vision Transformer are provided in Section II. In Section III, we will detail our proposed ST-MAE. Section IV presents the extensive experiments and associated analyses. More extended experiments and analysis are exhibited in Section V. Finally, conclusions and discussions will be given in Section VI.

\begin{figure*}[t]
\centerline{\includegraphics[width=\textwidth]{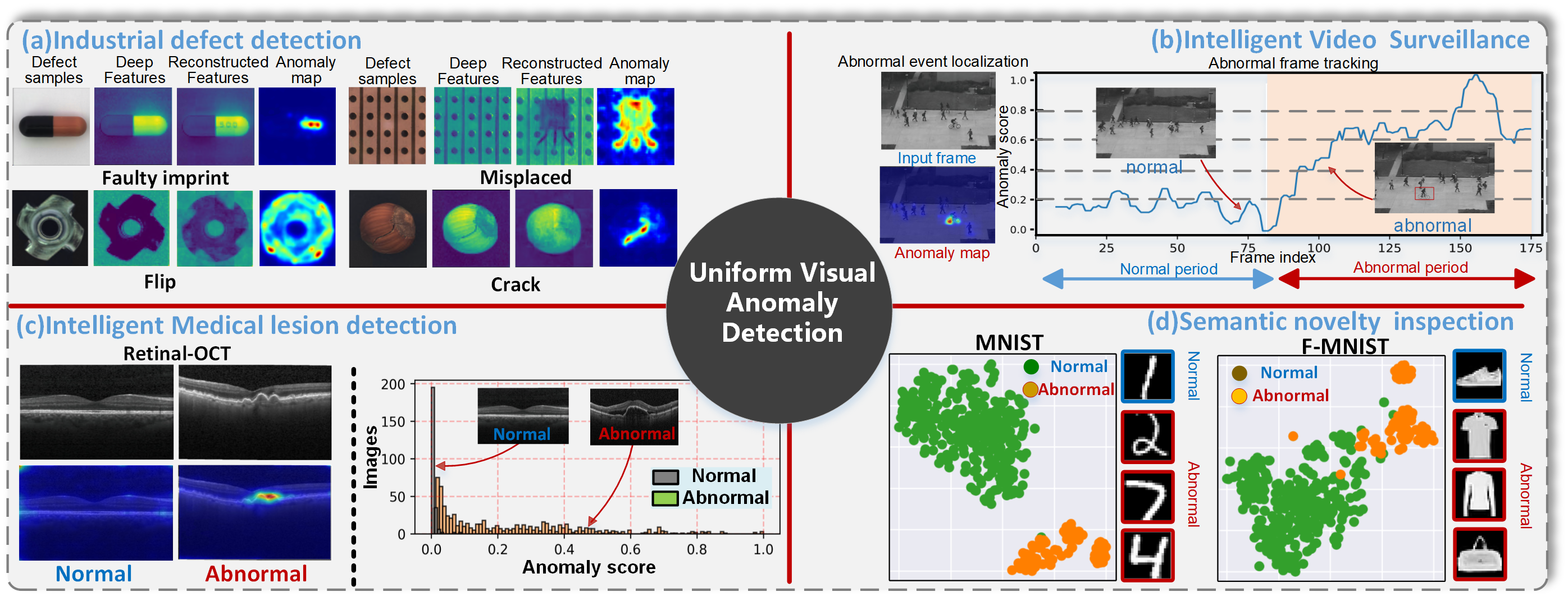}}
\caption[width=\textwidth]{
UVAD-oriented visual anomaly detection tasks in multiple application scenarios and the main features of our proposed ST-MAE framework. (a) Industrial defect detection task, the samples are sourced from the Mvtec AD dataset, and the selected defect types are regarded as difficult and unsolved in \cite{r64}. The deep features are visualized by selecting the values
from one channel. (b) Anomaly events detection task in the video. Our framework can simultaneously locate anomalous events within frame and track them at the inter-frame level. (c) Our ST-MAE can also handle the lesion detection task in intelligent medical imaging, the lesion samples are assigned a higher abnormality score (d) The one-class semantic novelty inspection. We illustrated that ST-MAE can distinguish novelty samples through T-SNE analysis of the latent representations extracted by ST-MAE.
}
\label{fig2}
\end{figure*}
\section{Related Work}

\subsection{Visual anomaly inspection approaches}
Anomaly inspection, also known as one-class novelty detection, is an active field in the community of machine learning and visual analysis. With the semantic shift, anomalies are defined as being outside the normal distribution. A natural option for handling this problem is to construct a discriminative boundary between the normal and abnormal samples. The one-class SVM (OC-SVM)\cite{r35} and the OC-SVDD\cite{r36} are proposed by learning a hyperplane or hypersphere surrounding the normal data and treating deviated samples as anomalies. Besides, another popular approach is unsupervised clustering, such as the Gaussian Mixture Models (GMM)\cite{r37} and the $k$-means algorithm. These methods are widely used but obtain suboptimal performance in applications where the data are high-dimensional due to the curse of dimensionality.

\subsection{Visual anomaly localization methods}
Unsupervised visual anomaly localization has attracted growing attention due to the increasing demands for more precise automated anomaly detection and has been extensively studied in a variety of genres. The mainstream approaches can be mainly categorized into two types: reconstruction-based and embedding-based models.

\subsubsection{Embedding-based Models}
Embedding-based models, essentially feature-distance-based models, are committed to detecting anomalous via the embedding distance between the samples in the discriminative high-dimensional feature space, where the abnormal images are embedded away from the normal feature clusters. Benefiting from the high semantic feature extraction, this kind of approach has been proven to be efficient for pixel-level anomaly segmentation. Recently, leveraging self-supervised learning\cite{r40}, some methods use the data augmentation\cite{r40}\cite{r41} on normal samples to build proxy tasks\cite{r9} to obtain robust normality representations. \cite{r9} proposed a Patch-SVDD framework that enriches the positional semantic information of embedding by leveraging jigsaw puzzle task. The SPADE\cite{r4} directly utilizes the deep embedded features distance as the anomaly assessment. The above two reveal effective performances on the benchmark Mvtec AD\cite{r21}, but the feature retrieval operation is needed during testing which is time-consuming and memory-intensive. PaDiM framework introduced in \cite{r11} and GCPF proposed in \cite{r5} relies on Gaussian distribution for normal reference and Mahalanobis distance for anomaly score, obtaining a relatively preferable testing efficiency. The student-teacher frameworks\cite{r10}\cite{r42} leverage the knowledge distillation to optimize the student networks from the knowledge of the pre-trained teacher network, and in the inference phase, the regression error is adopted as the anomaly score. Unfortunately, the dense sliding window scheme\cite{r10} is still inefficient in inference.
To sum up, the embedding methods yield reasonable results but are inferior in terms of the computation time for practical applications. 

\begin{figure*}[t]
\centerline{\includegraphics[width=\textwidth]{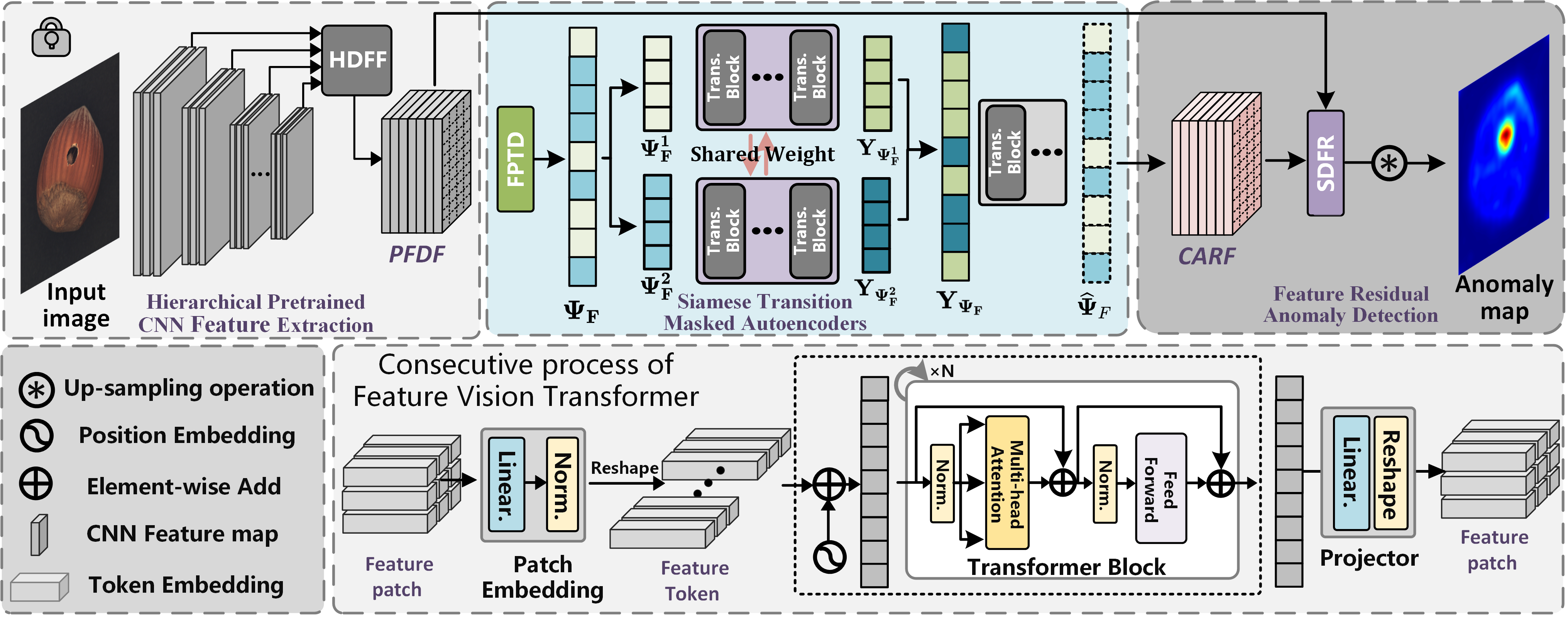}}
\caption[width=\textwidth]{
The pipeline of the ST-MAE framework. The overall structure is a DCNN-Transformer hybrid framework, which is constituted of three main modules: the LPSR, ST-MAE model, and SDFR. Note that our ST-MAE is a feature vision Transformer that operates on the deep features of DCNN, which is slightly different from the base ViT\cite{r61}, its consecutive computational process is roughly demonstrated.}
\label{fig2}
\end{figure*}

\subsubsection{Reconstruction-Based Models}

On the contrary, the reconstruction-based methods, which typically use the AE network architectures, are much more efficient, and the reconstruction error-based anomaly judgment is also computationally friendly, demonstrating greater potentiality for practical use. To settle the limitations of vanilla AE, several extended methods are developed. For instance, an MSCDAE\cite{r43} model is proposed to localize textural defects using a three-layer pyramid structure. MS-FCAE, which focuses on constraining the feature distribution for anomaly elimination, is also introduced in \cite{r44}. The structural loss-driven AE(AE-SSIM\cite{r6}) is proposed for the local structure restoration. Similarly, the generative adversarial network(GAN) is integrated with AE in \cite{r8}\cite{r45} to improve the reconstruction ability. For consideration of semantic information, the perceptual loss\cite{r46} is also proven to be effective for VAD both in the training and testing phases. The memory-augmented AE\cite{r13} that explicitly stores the normal pattern is proposed to mitigate the drawback of over-generalizability. The RIAD in \cite{r7} randomly drops partial patches of the image to convert the reconstruction to an inpainting problem, which is beneficial to anomaly elimination. Unlike the above-mentioned image reconstruction models, the Deep feature reconstruction(DFR)\cite{r14} detects anomalies based on feature reconstruction error.
In summary, most of the reconstruction models are still inefficient in terms of detection accuracy and are scenario-specific that can not be applied to different tasks simultaneously. 

\subsection{Recent advancement in vision Transformer}
Transformer architectures\cite{r15} have been widely investigated in natural language processing(NLP) and has been explored in the visual field recently. Vision Transformer(ViT)\cite{r16} is firstly proposed to be applied in large-scale image recognition. Many researchers have since employed it for a variety of downstream vision tasks. For instance, the DETR\cite{r17} model is proposed for object detection. From the sequence-to-sequence perspective, SETR\cite{r18} is also introduced for semantics segmentation. The swim Transformer proposed in \cite{r19} demonstrates great potential in various vision tasks. Furthermore, \cite{r20} proposed the Intra model that takes advantage of the vision Transformer and image inpainting for VAD, yielding effective performances on the benchmark Mvtec AD\cite{r21}. The Masked Autoencoder(MAE) proposed in \cite{r61} shows that the ViT can learn meaningful visual representations from the small proportion of visible patches subset, which yields promising performance in the downstream tasks. 

Combining the benefits of global and local context extraction, the ST-MAE proposed in the paper is a hybrid framework that takes full advantage of the Deep Convolutional Neural Networks(DCNN) and vision Transformer to learn the semantic normal pattern for the UVAD system. The ST-MAE utilizes the prep-trained DCNN as the feature extractor to convert low-level image pixels into discriminative deep features as property representation, then the features are randomly decoupled into two subsets in patch-wise, where the features can be reconstructed in the self-supervisory task of subsets transition. Finally, the task of anomaly detection can be solved by the feature reconstruction residual. Compared with the previous methods, the ST-MAE shows significantly superior performance and efficiency in a wide range of application scenarios.
\section{Proposed Methodology}

\subsection{Overall pipeline}
The overall pipeline of the proposed framework ST-MAE is presented in Fig. 4. Our ST-MAE is a holistic autoencoding framework that reconstructs the original signal. Unlike classical autoencoders, we adopt a hybrid design that integrates a deep convolutional neural network and a vision Transformer. The whole architecture of ST-MAE consists of three major components: Local perceptual semantics representation (LPSR) module, Siamese transition masked autoencoder(ST-MAE), and semantic deep feature residual(SDFR) module. As illustrated in Fig. 4, in the LPSR module, the given image as $I$ is firstly propagated into a pre-trained DCNN for extracting a pyramid-perceptual fusion dense feature(PFDF). Then, the PFDF is fed into the proposed ST-MAE model and is reconstructed as the Context-Aware Reassemble Feature(CARF). Finally, the SDFR module is employed to detect the anomalous properties and obtain the anomaly map utilizing the reconstruction residual of PFDF and CARF. In the subsections that follow, we go into detail about our methodology.

\subsection{Local perceptual semantics representation}
The current common reconstruction approaches are conducted within the image domain, this simplistic strategy of employing low-level representations of the images directly has troubles with the accurate reconstruction of image details, and more crucially, pixel-level representations are not sufficiently discriminative due to the limitations of their representational capabilities. Utilizing the intuition of embedding-based methods, for exploiting the semantic representation of the image, we proposed the LPSR, to fully leverages the prior knowledge of the pre-trained DCNN. Specifically, as depicted in Fig. 5, the LPSR consists of two sub-steps, pre-trained hierarchical feature extraction, and deep feature fusion. First, in the pre-trained hierarchical feature extraction step, the DCNN pre-trained on the ImageNet\cite{r27} is used as the embedded semantic feature extractor $\mathcal{F}$. Given the input image $I$, the outputs of different convolutional blocks of $\mathcal{F}$ are employed as the extracted hierarchical features:
\begin{equation}
\begin{aligned}
\psi_{F} &=\mathcal{F}\left ( I \right )\\
\psi_{F}&=\left \{\psi_{F}^{1}, \psi_{F}^{2},...,\psi_{F}^{\mathcal{H}}  \right \} 
\end{aligned}
\end{equation}
where the $\psi_{F}^{i}\in \mathbb{R} ^{H^{i}\times W^{i}\times C^{i}}$ denotes the $i$-th hierarchical feature with the size of $H^{i}\times W^{i}\times C^{i}$ and $\mathcal{H}$ is the number of the hierarchies. The hierarchical features of the image are extensively extracted by the high-dimensional mapping operations, which have been equipped with sufficient discriminability through the integration of contextual semantics. Moreover, the idea of multi-scale aggregation is also introduced to prompt the representation capability. In general, the low-level features of the small receptive field are used as simple edge or texture descriptors, and with the network level deepens, the feature maps corresponding to the large receptive fields are more semantic. Thus, benefiting from rich receptive information, fusing multi-scale features will promote representation performance. In the deep feature fusion, different feature maps are interpolated into the same spatial scale and concatenated in the depth dimension:
\begin{equation}
\mathbf{\Phi _{F}} =\left \{ {\textstyle \bigcup_{i=1}^{\mathcal{H}}}  \Gamma \left ( \psi_{F}^{i} \right )\mid i\in \left \{ 1,2,...,\mathcal{H}  \right \} \right \}
\end{equation}
where the $\mathbf{\Phi _{F}}\in \mathbb{R} ^{H\times W\times C}$ represents the pyramid-perceptual fusion dense feature(PFDF), ${\bigcup}$ is the channel-wise concatenation, and $\Gamma$ denotes the interpolation operation. 

\begin{figure}[t]\centering
\includegraphics[width=8.8cm]{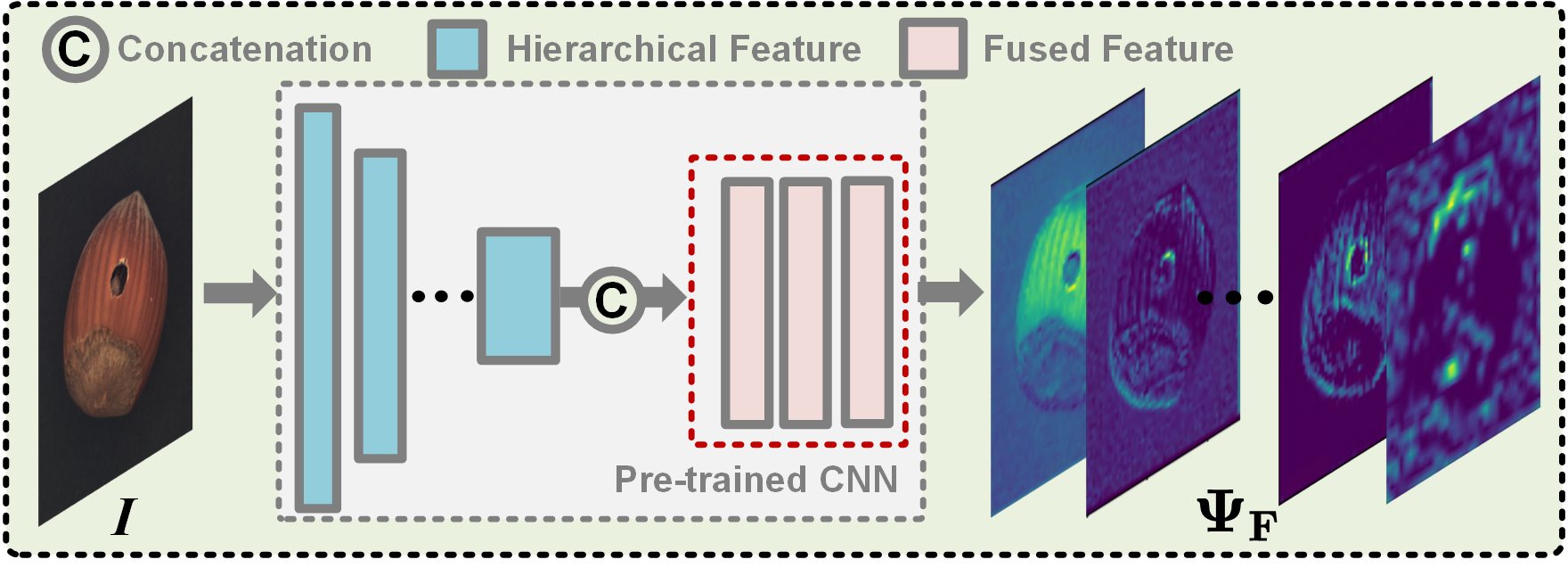}
\caption{LPSR schematic. LPSR is used to extract and fuse the latent multi-level semantic features with the pre-trained DCNN. The hierarchical feature maps are firstly resized into the same spatial size and then fused.}
\label{FIG_3}
\end{figure}
In short, compared with image reconstruction-based approaches that employ the image pixel as representation, the proposed LPSR module constructs multi-scale discriminative deep features as property descriptors through the pre-trained hierarchical feature extraction and deep feature fusion to improve the efficiency of spatial-aware contextual semantics.

\subsection{Siamese transition masked autoencoder}
\subsubsection{Feature patches token decoupling}
Following ViT\cite{r16}, we divide the PFDF into regular non-overlapping feature patches. As shown in Fig. 6, the deep feature PFDF $\mathbf{\Phi_{F}}$ can be viewed as a set of feature patches, which contains the local structural context of the input image. To divide these feature patches, in particular, let $K$ be the desired size of the feature patch and $N=\frac{H}{K} \times \frac{W}{K}$ is the total number of feature patches. The PFDF is altered by first partitioning the feature map into square regions of size $K$ and then splitting it into a $N$ grid of square patches. Then, just as in a standard ViT, the square patches are reshaped into flattened feature patch tokens, which are additionally mapped to $D$ dimensions with a learnable linear projection and added to the corresponding position embeddings intending to retain the spatial positional information:
\begin{equation}
\mathbf{\Psi}_{F}= [\mathcal{X} _{\Phi_{F}}^{1,1}\mathbf{E}, \mathcal{X}_{\Phi_{F}}^{i,j}\mathbf{E},...,\mathcal{X}_{\Phi_{F}}^{\sqrt{N} ,\sqrt{N}}\mathbf{E}]+\mathbf{E}_{pos}
\end{equation}
where the $\mathcal{X} _{\Phi _{F}}^{i,j}\in \mathbb{R} ^{  K^{2} \times C}$ is the flattened feature patch with the position of the $i$-th row and $j$-th column. $\mathbf{\Psi}_{F}$ is the feature patch embedding sequence of the $\mathbf{\Phi _{F}}$, $\mathbf{E}  \in \mathbb{R} ^{\left ( K^{2}\cdot  C \right ) \times D}$ is the trainable projection matrix and $\mathbf{E}_{pos}$ refers the positional embeddings. Subsequently, the feature patches token decoupling (FPTD) is constructed to split the full set $\mathbf{\Psi}_{F}$ into two disjoint subsets. Our decoupling strategy is straightforward: each patch embedding is randomly sampled following a uniform distribution and the unsampled ones are masked, dividing the full set of sequence into two disjoint subset sets $\mathbf{\Psi}_{F}^{1}$ and  $\mathbf{\Psi}_{F}^{2}$. Each contains half the number of all embedding tokens, and the entire sequence is as the union of the tokens belonging to subsets:

\begin{equation}
\left \{ \mathbf{\Psi}_{F}^{1},\mathbf{\Psi}_{F}^{2}\mid \mathbf{\Psi}_{F}^{p}\in \mathbb{R}^{N/2\times D}, p\in\left \{ 1,2 \right \} \right \}\\
\end{equation}

\begin{figure}[t]\centering
\includegraphics[width=8.8cm]{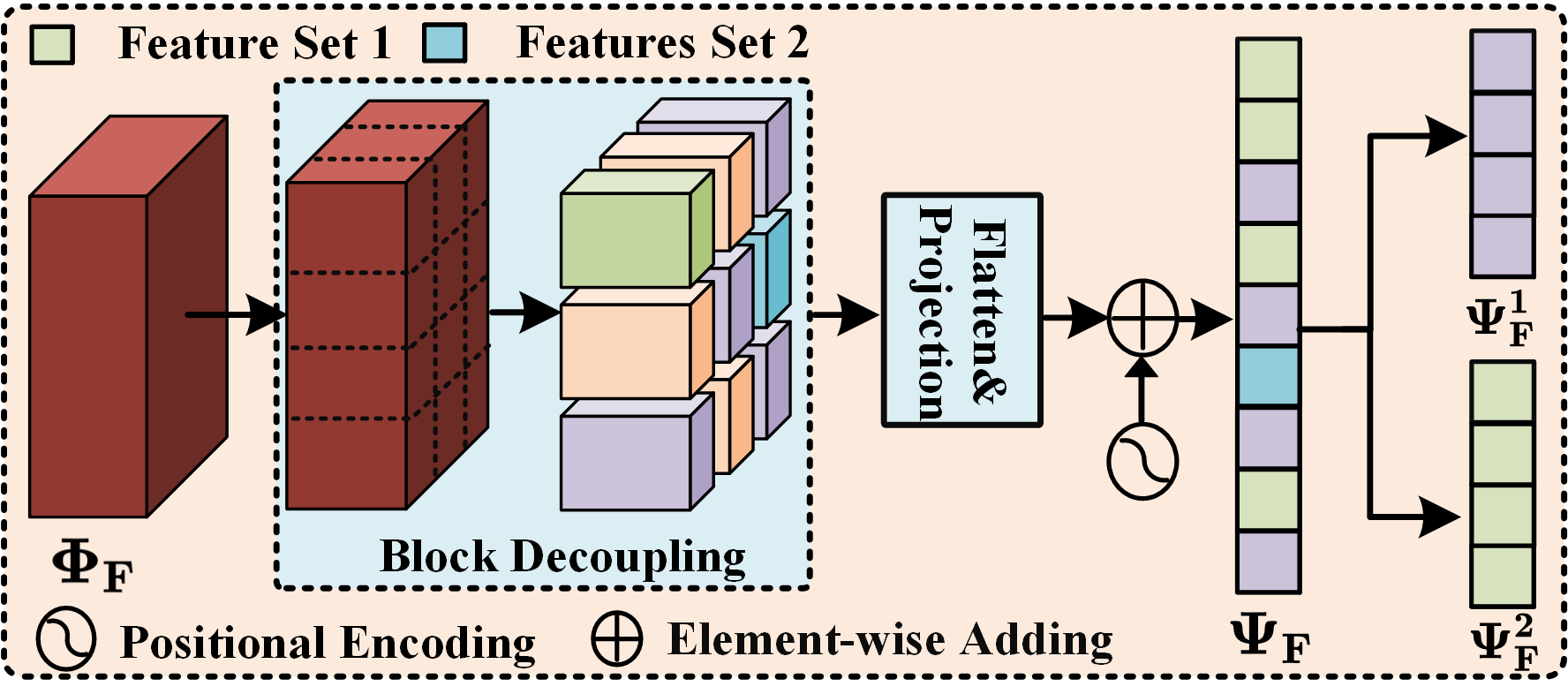}
\caption{The architecture of the FBTD. The CNN feature map can be viewed as a set of feature patches. First, the PFDF are decoupled into two disjoint feature sets in patch-wise. then the decoupled feature patches are embedded into token sequences that are fed into the Transformer model.}
\label{FIG_3}
\end{figure}

By applying the FPTD, two complementary feature patch embedding sequence subsets are constructed with PFDF. Then, the Siamese transition masked Transformer(ST-MAE) model is employed to reconstruct the full set based on the proposed heuristic deep feature transition paradigm, described next.

\subsubsection{Deep feature transition}

In this study, a simple stack of Transformer blocks is adopted as the network structure of the ST-MAE. As shown in Fig. 4, each block is composed of a multi-head attention block(MAB) and feed-forward block (FFB) which contains two hidden layers with GELU non-linearity. For each block,  Layernorm (LN) is applied and the residual connection is employed, the consecutive computational process of transformer blocks is denoted as\cite{r16}:
\begin{equation}
\begin{aligned}
\mathbf{Z}_{0} &=\mathcal{LN}\big( \mathbf{\Psi}_{F}^{p} \big) \\
{\mathbf{Z}}'_{\ell} & = \mathcal{MAB} \left ( \mathcal{LN}\left ( \mathbf{Z}_{\ell-1} \right )  \right ) +\mathbf{Z}_{\ell-1} &\quad & \ell =\left \{  1,...,L \right \}\\
\mathbf{Z}_{\ell} & = \mathcal{FFB} \left ( \mathcal{LN}\left ( {\mathbf{Z}}'_{\ell} \right )  \right ) +{\mathbf{Z}}'_{\ell}&\quad & \ell =\left \{  1,...,L \right \}\\
\end{aligned}
\end{equation}
 where ${\mathbf{Z}}'_{\ell}$ and $\mathbf{Z}_{\ell} $ denote the output of the MAB and the FFB for block $\ell$, respectively.

Different from the existing typical vision transformers, our ST-MAE adopts a Siamese encoder structure which is applied to the visible, unmasked feature patch embeddings of two disjoint subsets to encode them into the latent space as $Y_{\mathbf{\Psi}_{F}^{1}}$ and $Y_{\mathbf{\Psi}_{F}^{2}}$. Like MAE \cite{r61}, each encoder in our approach maps the partially observed signal to the latent representation, which has been approved to be effective to learn object semantics from images. Nevertheless, using the two disjoint subsets as described before, each encoder in the proposed Siamese structure operates on one of the subsets, which guarantees that no information loss of the original signal. After encoding, we reassemble two subsets of latent representations by the proposed deep feature transition.

As mentioned before, the feature patch embeddings of the full set are divided into two subsets, each embedding is associated with a position location index that indicates its location in the sequence of the full set. Thus, the entire sequence can be reconstructed from two subset sequences based on the location index lists of two subsets. In this manner, all the embeddings in the reconstructed sequence are aligned with themselves. As depicted in Fig. 7, instead of reassembling the encoded embedding subsets using this trivial solution, we presented the deep feature transition, where the two encoded embedding subsets are reassembled as the entire latent embedding sequence of the full set by using the position index list of their complementary subset. More essentially, the two latent embedding subsets are equivalent to being mutually transitioned in the reassembly process. By performing the deep feature transition in the latent space, an implicit self-supervisory signal is derived. This can be explained by that in the reassembled latent embedding sequence of the full set, the embeddings of each subset are generated by another complementary subset, which is masked and invisible 
in the encoding phase. Thus, this strategy encourages the model to learn the useful features of a holistic understanding of object semantics.

Then, positional embeddings are added to all latent embeddings in the reassembled entire sequence of the full set, which are subsequently handled by a lightweight decoder(Consisting of another serial of Transformer blocks). Finally, as shown on Fig. 4, the output tokens of the decoder in $ \mathbb{R} ^{D}$ space are mapped back to the space of the flattened feature patch $ \mathbb{R} ^{K^{2}\cdot C}$ via a learnable projection head. The CARF $\hat{\mathbf{\Phi}}_{F}\in \mathbb{R} ^{H\times W\times C}$ is obtained by resizing the $\hat{\mathbf{\Psi}}_{F}$ back to 2-D feature map.

\begin{figure}[t]\centering
\includegraphics[width=8.8cm]{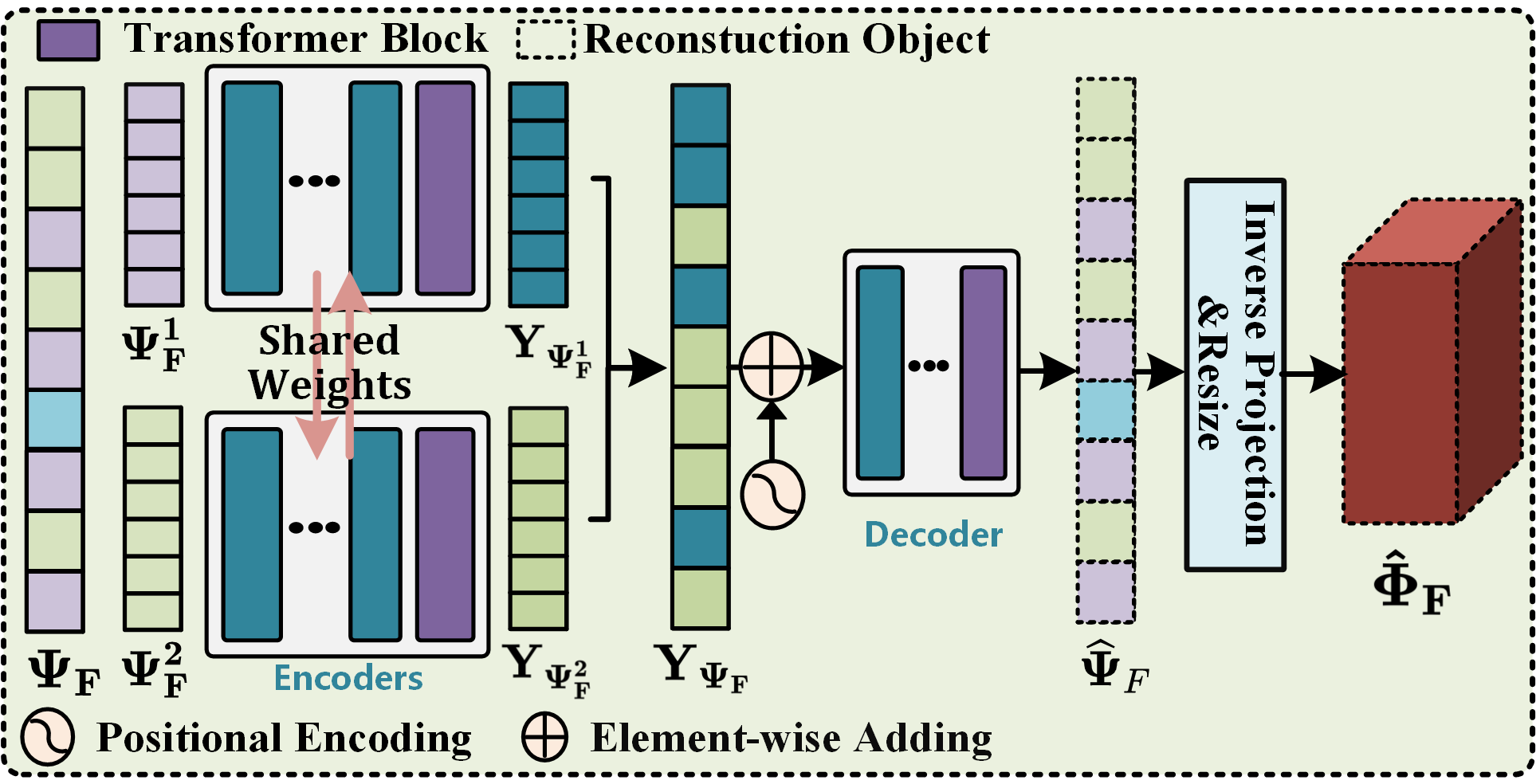}
\caption{Illustration of Siamese transition masked Transformer. The two decoupled feature patch token sequences are fed into the Siamese encoder and perform the mutual transition when reassembled in the latent space. The reconstructed full set of feature patch token sequences is then obtained by the decoder, which finally forms the CARF.}
\label{FIG_3}
\end{figure}
Notably, in contrast to reconstruction approaches, each subset $\mathbf{\Psi}_{F}^{p}$ is transitioned to the complementary parts that do not belong to itself in the latent space, which guarantees that each embedding tokens in the $Y_{\mathbf{\Psi}_{F}}$ are generated by conditioning only on the information of their complementary subset and thus semantically plausible, avoiding the trivial anomaly reconstruction observed in previous approaches, which is very beneficial to VAD.

\subsection{Semantic deep feature residual}
Instead of reconstructing the images in a pixel-wise manner like the MAE\cite{r61} and Intr\cite{r20}, we adopted the feature-level measurement, the relaxed version of the pixel-level constraints, for better robustness.
\subsubsection{Objective function for training phase}
we measure the distance between the feature patches for normal images with two types of losses. The losses include intensity loss and orientational loss. For intensity loss, this loss function aims to minimize the Euclidean distance of the activation values between the input patches and the reconstructed patches. Thus, the intensity loss $\mathcal{L}_{int} $ is formulated as:
\begin{equation}
\mathcal{L}_{int}\big(\mathbf{\hat{\Phi} _{F}},\mathbf{\Phi_{F}} \big) = \sum_{i=1}^{\sqrt{N}} \sum_{j=1}^{\sqrt{N} }{\left \| \mathcal{X}_{\hat{\Phi}_{F}}^{i,j}-\mathcal{X} _{{\Phi_{F}}}^{i,j} \right \|^{2}_{2}} 
\end{equation}
where $\left \| \cdot  \right \|_{2}$ represents the $\ell 2$ norm. Additionally, due to the 
high-dimensional nature of the deep features, we further employ the orientational loss to ensure the homo-directional of the deep feature vectors:
\begin{equation}
\mathcal{L}_{ori}\big(\mathbf{\hat{\Phi} _{F}},\mathbf{\Phi_{F}} \big) = \sum_{i=1}^{\sqrt{N}} \sum_{j=1}^{\sqrt{N} }  1-\frac{\big( \mathcal{X}_{\hat{\Phi}_{F}}^{i,j}  \big)^{T}\cdot \big( \mathcal{X} _{{\Phi_{F}}}^{i,j}\big)}{\left \| \mathcal{X}_{\hat{\Phi}_{F}}^{i,j}  \right \|\cdot \left \| \mathcal{X} _{{\Phi_{F}}}^{i,j}  \right \|} 
\end{equation}
Considering the above two types of loss, the total loss function for ST-MAE's optimization is formulated as follows:
\begin{equation}
\mathcal{L}\big(  \mathbf{\hat{\Phi} _{F}},\mathbf{\Phi_{F}} \ \big) = \mathcal{L}_{int}\big(  \mathbf{\hat{\Phi} _{F}},\mathbf{\Phi_{F}} \big)+\lambda \mathcal{L}_{ori}\big(  \mathbf{\hat{\Phi} _{F}},\mathbf{\Phi_{F}} \ \big)
\end{equation}
where $\lambda$ is the weight parameter that balances the relative contributions of the two types of losses and is set to 5 in practice.

\subsubsection{Anomaly score for testing phase}
In the testing phase, the whole system can be deployed for online anomaly detection. When the input query sample is anomalous, the ST-MAE is likely to fail in abnormal latent feature transition, since the ST-MAE only learns to transition anomaly-free latent representations. Thus, ST-MAE is capable of reflecting abnormality using the feature residual of PFDF and CARF. Consistent with the training phase, the feature residual can be measured by the intensity divergence and orientational divergence, denoted as
\begin{equation}
\begin{aligned}
\mathcal{A}_{i}(h,w) & = \left \| \mathbf{\hat{\Phi} _{F}}(h,w)- \mathbf{\Phi _{F}}(h,w)\right \|^{2}_{2} \\
\mathcal{A}_{o}(h,w) & = 1-\frac{\big( \mathbf{\hat{\Phi} _{F}}(h,w) \big)^{T}\cdot\big(  \mathbf{\Phi _{F}}(h,w) \big)}{\big\|  \mathbf{\hat{\Phi} _{F}}(h,w) \big\| \cdot \big\|  \mathbf{\Phi _{F}}(h,w)  \big\| } 
\end{aligned}
\end{equation}
where the $\mathcal{A}_{int}(h,w)$ and $\mathcal{A}_{ori}(h,w)$ represent the anomaly score map of the intensity and orientational discrepancy activation responses at the position of $(h,w)$. Then two anomaly score maps are fused for the final anomaly map $\mathcal{A}$ in a multiply manner.

Finally, the interpolation operation is employed to change the spatial resolution of the anomaly map to the input image size. A Gaussian filter($\sigma =4$) is applied to smooth the final anomaly map. As suggested in \cite{r12}, the standard deviation of $\mathcal{A}$ is utilized as the image-level anomaly score.

\begin{table*}
\caption{{Image-level/pixel-level AUC ROC results of different methods in MVTec AD}}
\label{table}
\setlength{\tabcolsep}{3pt}
\begin{threeparttable}
\begin{tabular}{p{\textwidth}}
$\includegraphics[width=\textwidth]{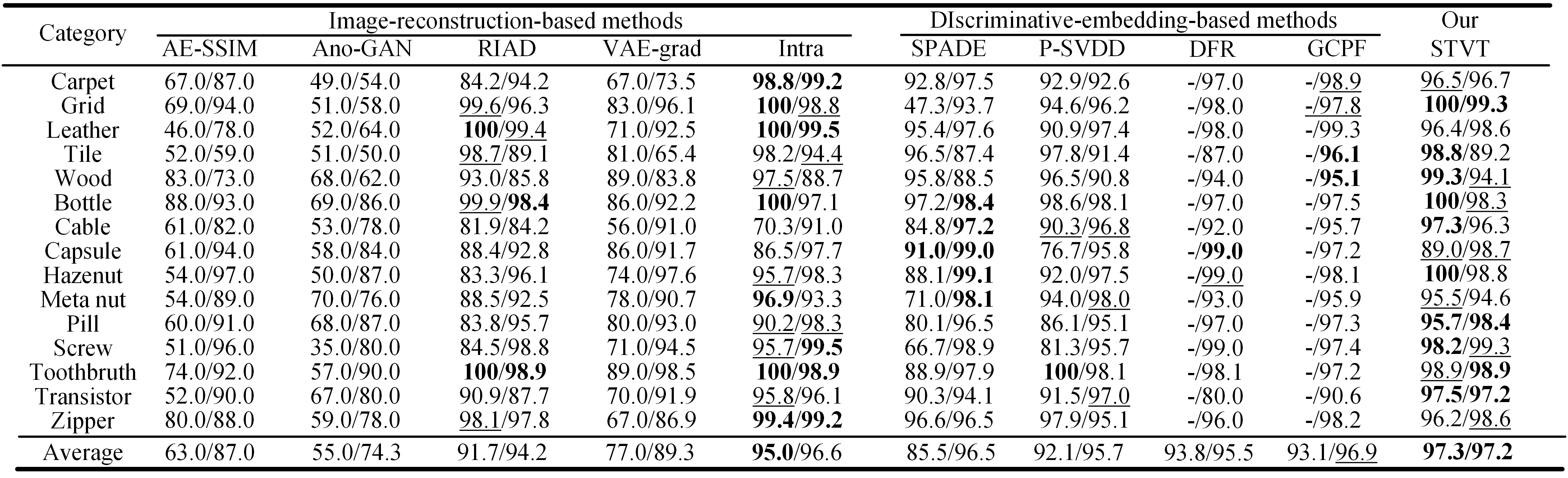}$
\end{tabular}
\begin{tablenotes}
       \footnotesize
       \item[1]The best AUC ROC performance is indicated by bold font, while the second best is indicated by an underline.
\end{tablenotes}
\end{threeparttable}
\label{table4}
\end{table*}

\begin{table}
\caption{Details of ST-MAE model variants.}
\label{table}
\setlength{\tabcolsep}{3pt}
\begin{tabular}{p{\columnwidth}}
$\includegraphics[width=\columnwidth]{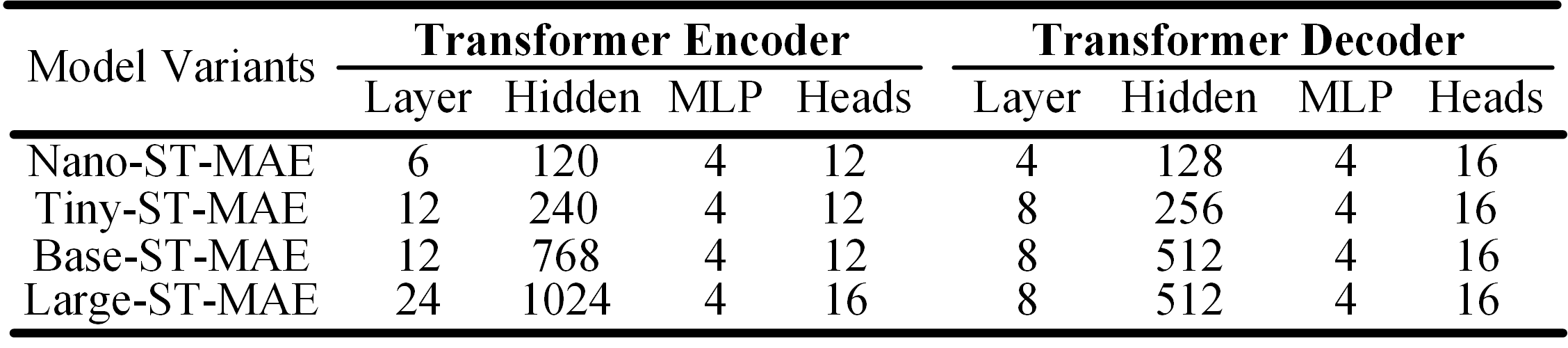}$
\end{tabular}
\label{table2}
\end{table}

\begin{figure*}[t]
\centerline{\includegraphics[width=\textwidth]{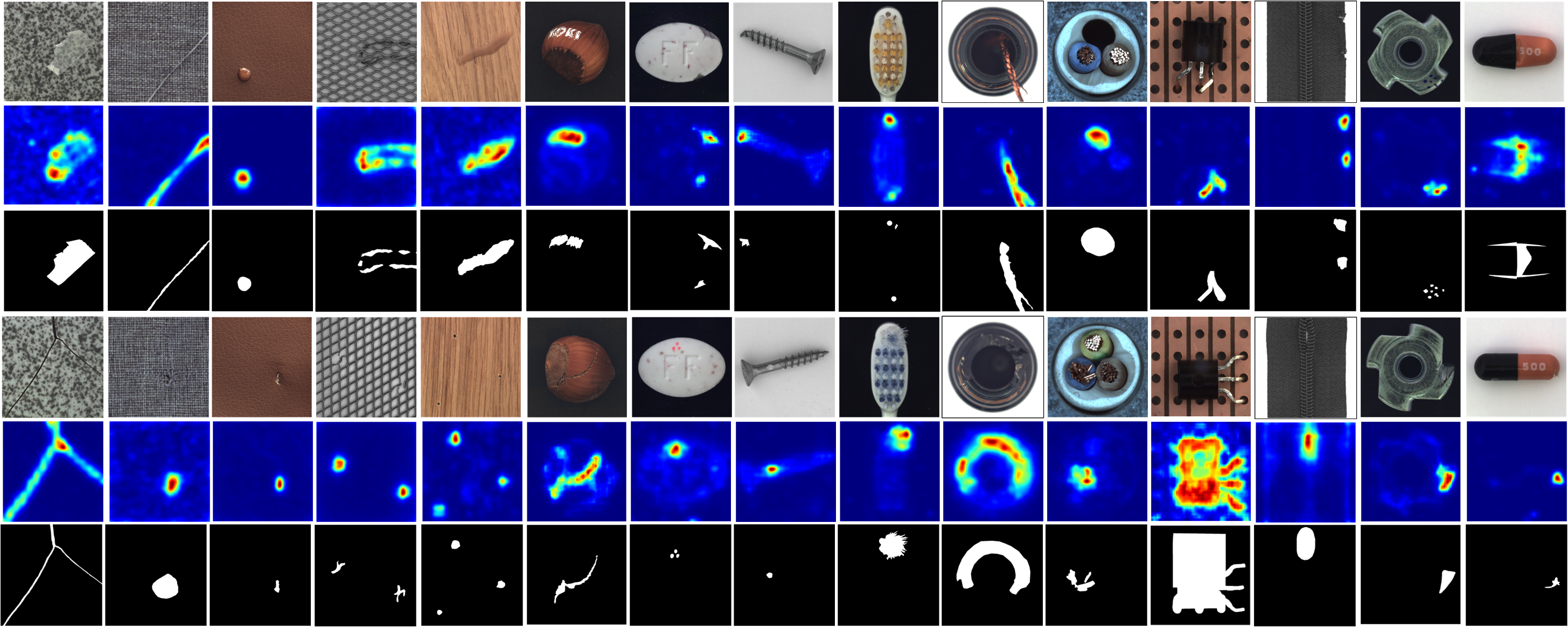}}
\caption[width=\textwidth]{
Qualitative results on the MVTec AD dataset. From top to bottom are shown the original defect samples, the inspection results obtained using the proposed ST-MAE method, and the ground truth.
}
\label{fig2}
\end{figure*}

\section{Experiment results}
In this section, we report several sets of experiments conducted on existing VAD benchmarks to analyze the effectiveness of the proposed method. While the ablation studies are also experimentally carried out and discussed.

\subsection{Implementation setup and evaluation metric} Although existing Transformer models like ViT\cite{r16} and MAE\cite{r61} are usually trained on the large-scale dataset, We trained the ST-MAE model on the limited samples from scratch by an AdamW optimizer with a learning rate of 1e-4 and batch size of 8 for 400 training epochs, while the weights of the LPSR are freeze during the training phase. For the LPSR module, the VGG19 network\cite{r26} pre-trained on the ImageNet dataset is exploited as the feature extractor by default, and the multi-scale features are extracted at the last ReLU layer of the first four convolutional blocks. For the configuration of the ST-MAE model, we validated different model variants as summarized in Table II. Without extra specifications, we adopt the base model. All the experiments are implemented on a computer with Xeon(R) Gold 6226R CPU@2.90GHZ and NVIDIA A100 GPU with 40GB memory size.

In our experiments, for the anomaly localization task in defect detection, video anomaly detection, and medical lesion detection, each image is resized into the resolution of 256$\times$256, and the pixel intensities are normalized according to the standard deviation and mean value obtained on the ImageNet dataset. The spatial size of PFDF is set to $64\times64$, the number of channels is the sum of that of all fused features which is equal to 960, and the hyper-parameter of the patch size $K$ in the FPTD module is set as 4 by default; for the semantic novelty inspection task, the image size is set as 32$\times$32, the spatial size of PFDF is set to $8\times8$, and the patch size $K$ is  set as 1.

Following the convention in prior works, the area under the receiver operating characteristic (AUC ROC) is adopted as the primary quantitative evaluation metric, and the accuracy, F1-score and average precision are additionally employed. For all criteria, a higher value indicates better performance.

\subsection{Datasets configuration}

\begin{table}
\caption{Pixel-level Localization AUC ROC comparison results on the DAGM and NanoTWICE datasets.}
\label{table}
\setlength{\tabcolsep}{3pt}
\begin{tabular}{p{\columnwidth}}
$\includegraphics[width=\columnwidth]{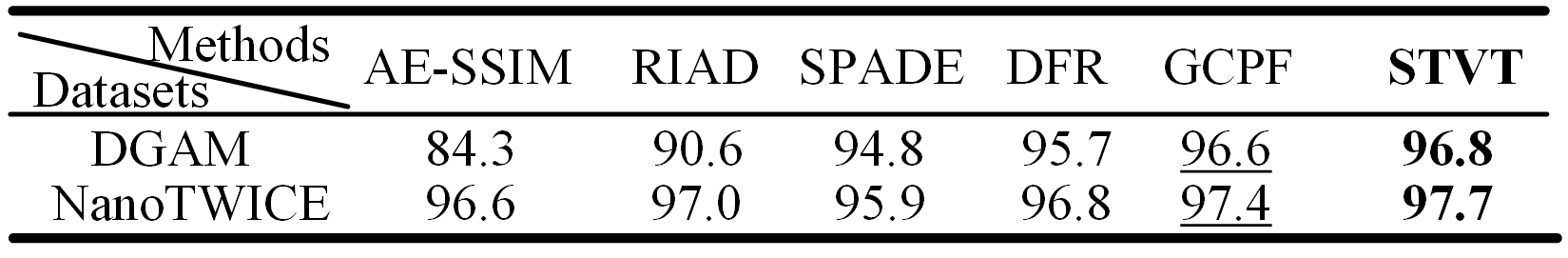}$
\end{tabular}
\label{table2}
\end{table}

All the experiments are conducted on a comprehensive dataset composed of multiple benchmarks including the industrial image samples, medical image samples and surveillance video samples as described below.

For industrial scenarios, the Mvtec AD\cite{r21} dataset, which has been widely investigated for unsupervised anomaly detection and localization recently, is adopted. With 5 classes of textural defects and 10 classes of object defects, it is comprised of 3629 normal images for training; 467 normal images, and 1258 abnormal images for testing. The defects are diverse and representative of real industrial scenarios, which is still a challenging task. In addition, the NanoTWICE\cite{r22} and DAGM\cite{r23} datasets are also employed for performance verification.
 
In terms of lesion detection in the medical image field, the Retinal-OCT dataset\cite{r29} is utilized. It is made up of 84,495 clinical samples soured from retinal optical coherence tomography (OCT) images. Three types of anomalies are included in the testing sets, including CNV, DME, and DRUSEN. Consistent with the original training-testing split settings, we only use the normal case samples in the training set for model training and use all testing images for performance validation.

To further verify the generality of ST-MAE, the surveillance video anomaly detection on UCSD Ped2 and CUHK Avenue datasets is also considered in this article. UCSD Ped2 consists of 4560 frames which are split into 16 training videos and 12 testing videos, respectively. The abnormal cases cover bicycles, skateboarders, and vehicles within the normal crowd. CUHK Avenue is composed of 37 videos, of which 16 are used for training and 21 are used for testing, respectively. The test video contains 47 anomalies such as running, throwing objects, and dancing.

\begin{table}
\caption{Image-level inspection performance comparison results on the Retinal-OCT dataset}
\label{table}
\setlength{\tabcolsep}{3pt}
\begin{tabular}{p{\columnwidth}}
$\includegraphics[width=\columnwidth]{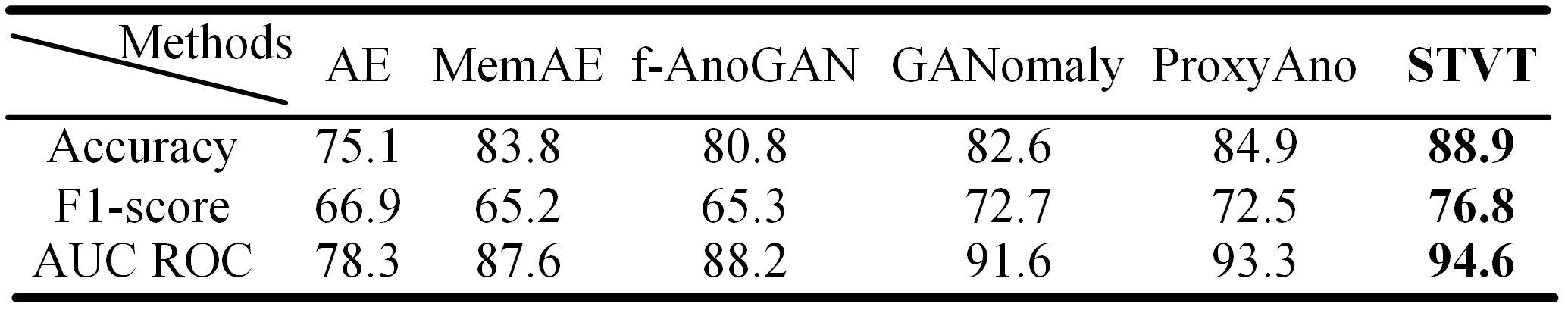}$
\end{tabular}
\label{table2}
\end{table}
In addition, our experiments also involved the one-class semantics datasets for the performance verification of the proposed ST-MAE method on the demand of UVAD, specifically referring to MNIST\cite{r60}, Fashion-MNIST\cite{r58} and CIFAR-10\cite{r59} three datasets. We follow the original setting to split the total dataset into training and testing sets. All three datasets are semantically divided into 10 sub-classes, only one sub-class is utilized in the training which is regarded as normal in the testing phase, while the other classes are all treated as abnormal.

\subsection{Overall performance evaluation}
\subsubsection{Case study 1: Industrial application scenarios}
In this subsection, the proposed ST-MAE is compared on Mvtec AD dataset with the long-standing image reconstruction based baselines AE-SSIM\cite{r6}, Ano-GAN\cite{r24}, RIAD\cite{r7}, VAE-grad\cite{r25}, Intra\cite{r20}, and the discriminative-embedding-based approaches SPADE\cite{r4}, P-SVDD\cite{r9}, DFR\cite{r14} and GCPF\cite{r5}. All the methods are applied for image-level and pixel-level VAD tasks.

\begin{figure*}[t]
\centerline{\includegraphics[width=\textwidth]{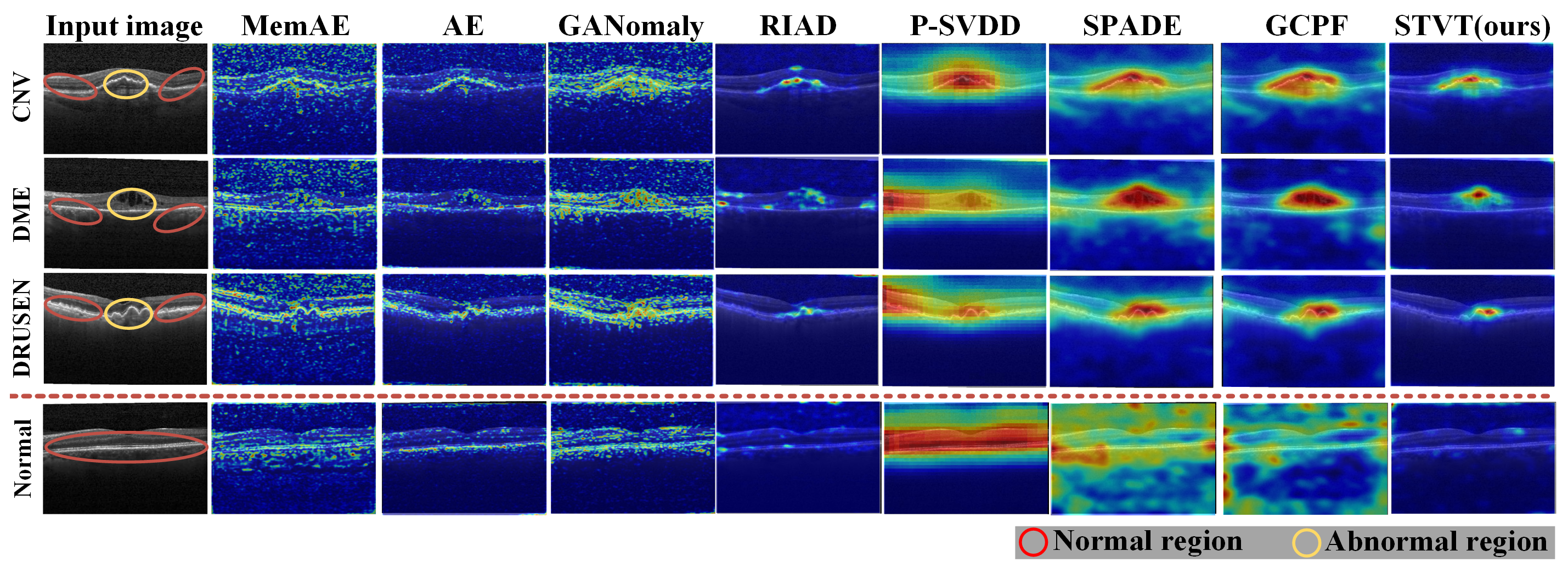}}
\caption[width=\textwidth]{
 Some examples of the medical lesion Localization results of the compared methods on the Retinal-OCT dataset, in which the abnormal and normal regions are identified by the yellow and red ellipses, respectively.}
\label{fig2}
\end{figure*}

\begin{table}
\caption{AUC ROC of different methods on Ped2 and Avenue datasets.}
\label{table}
\setlength{\tabcolsep}{3pt}
\begin{tabular}{p{\columnwidth}}
$\includegraphics[width=\columnwidth]{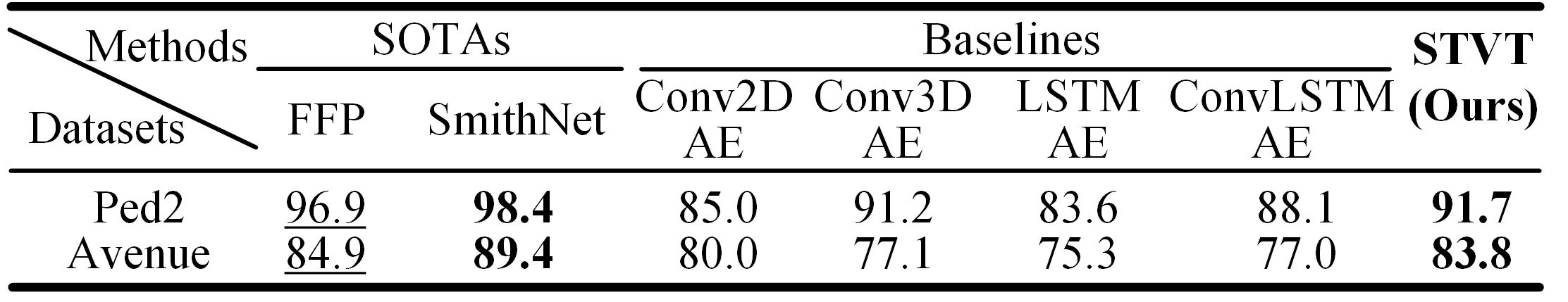}$
\end{tabular}
\label{table2}
\end{table} 

The comparison results of different methods are reported in TABLE I. It is noted that ST-MAE achieves the comprehensive 97.3/97.2 for image/pixel-level AUC ROC over all 15 categories. Among all the compared methods, ST-MAE achieves the best or the second-best accuracy on 9 categories of defects for pixel-wise localization and 13 categories of defects for image-wise inspection. Our method demonstrates robust performance for all categories datasets and only gets a slightly poor performance on the tile category. Moreover, the above-mentioned approaches in TABLE I are categorized into two groups, namely image-reconstruction-based methods and discriminative-embedding-based methods, the latter have better overall performance. Although our ST-MAE follows the paradigm of reconstruction, as evident from TABLE I, our method still outperformed the most recent cutting-edge methods for pixel-level localization and image-level inspection. Furthermore, for the most challenging category transistor, the ST-MAE can obtain 97.5$\%$/97.2$\%$ AUC ROC, substantially surpassing the other methods. All the experimental results mentioned above demonstrate the remarkable performance of the proposed ST-MAE.

To further verify the effectiveness of ST-MAE, we extended it to the pixel-level localization task of the NanoTWICE dataset and DAGM dataset. The performance of the ST-MAE is compared with the competing models, including the AE-SSIM\cite{r6}, RIAD\cite{r7}, SPADE\cite{r4}, DFR\cite{r14} and GCPF\cite{r5}. As reported in Table III, ST-MAE enhances the AUC ROC values by margins of 12.5, 6.2, 2, 1.1, 0.2 for the NanoTWICE dataset and 1.1, 0.7, 1.8, 0.9, and 0.3 for DAGM dataset when versus the baseline techniques, respectively. 

\begin{figure}[t]\centering
\includegraphics[width=8.8cm]{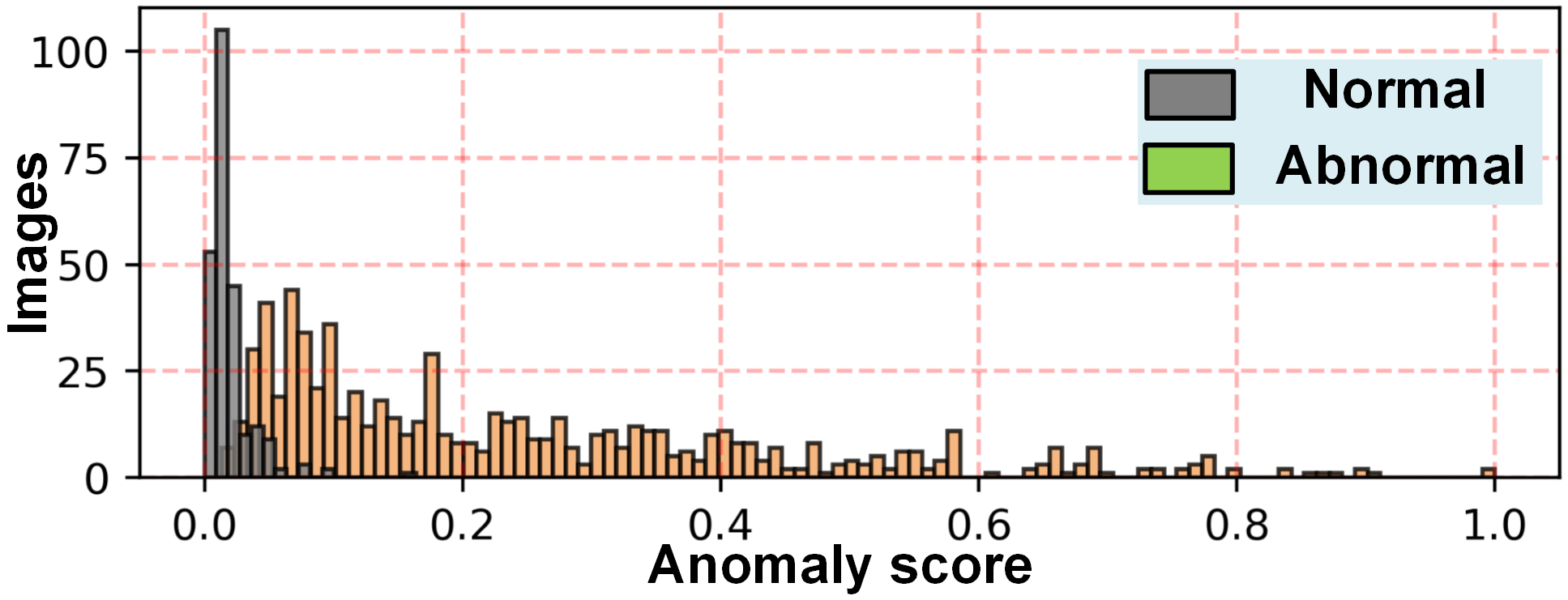}
\caption{Histogram of anomaly scores for the normal and abnormal samples in the Retinal-OCT dataset.}
\label{FIG_3}
\end{figure}

\begin{figure}[t]\centering
\includegraphics[width=8.8cm]{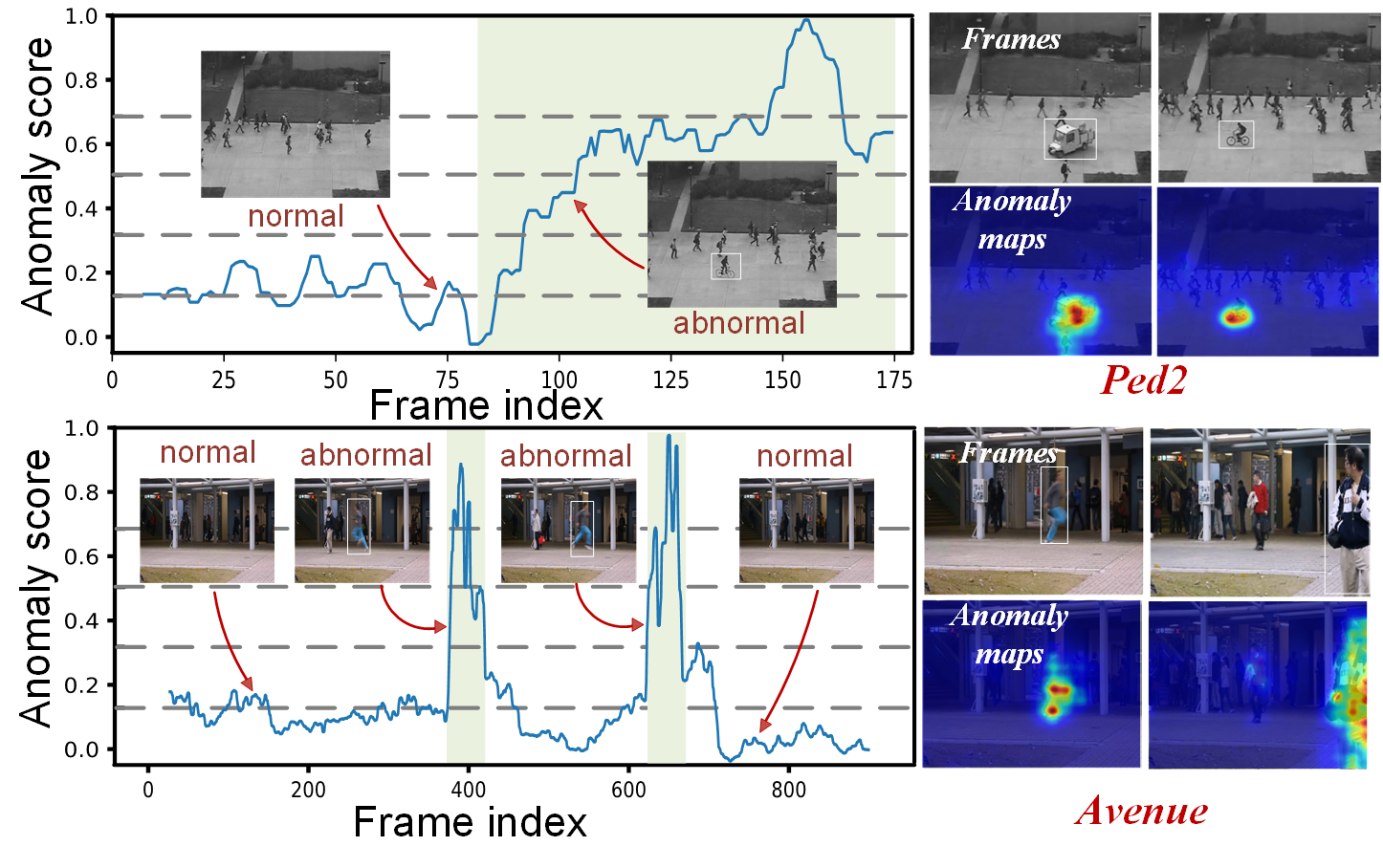}
\caption{Examples of detection performance of STVT on the Ped2 and Avenue datasets. The left side of the figure shows the anomaly score curves for two testing video instances, and the right side displays the anomaly maps of some testing frames.}
\label{FIG_3}
\end{figure}
Fig. 8 presents the qualitative results of ST-MAE on the MVTec AD dataset, which reveals its remarkable capability of simultaneous localization for various defect categories.

\subsubsection{Case study 2: Medical lesion detection} As stated before, the proposed ST-MAE is suitable for a wide range of application scenarios. Therefore, the ST-MAE model is validated in practice for medical image anomaly detection. As shown in Table. IV, the AUC ROC comparison results of image level anomaly inspection on the Retinal-OCT dataset are presented. The results reveal that the proposed STVT significantly outperforms the long-standing baseline methods AE\cite{r55}, MemAE\cite{r13}, f-AnoGAN\cite{r56}, and GANomaly\cite{r57}. The cutting-edge method ProxyAno\cite{r54}, which is specifically designed for detecting an anomaly in medical images, takes advantage of large amounts of additional data for model training. In detail, ProxyAno uses extra proxy images and pseudo-abnormal images, which are generated by laborious human design methods. In contrast, without any extra training data, the proposed STVT method only uses the original retinal images in the training phase, but still achieves a better result with the image level AUC ROC of 94.6, which is superior to ProxyAno(93.3). Moreover, the STVT improves the accuracy and F1-score score by 4.0 and 4.3 compared with the ProxyAno. The results strongly demonstrate the effectiveness of our proposed STVT for the unsupervised medical lesion anomaly detection task.

\begin{figure*}[t]
\centerline{\includegraphics[width=\textwidth]{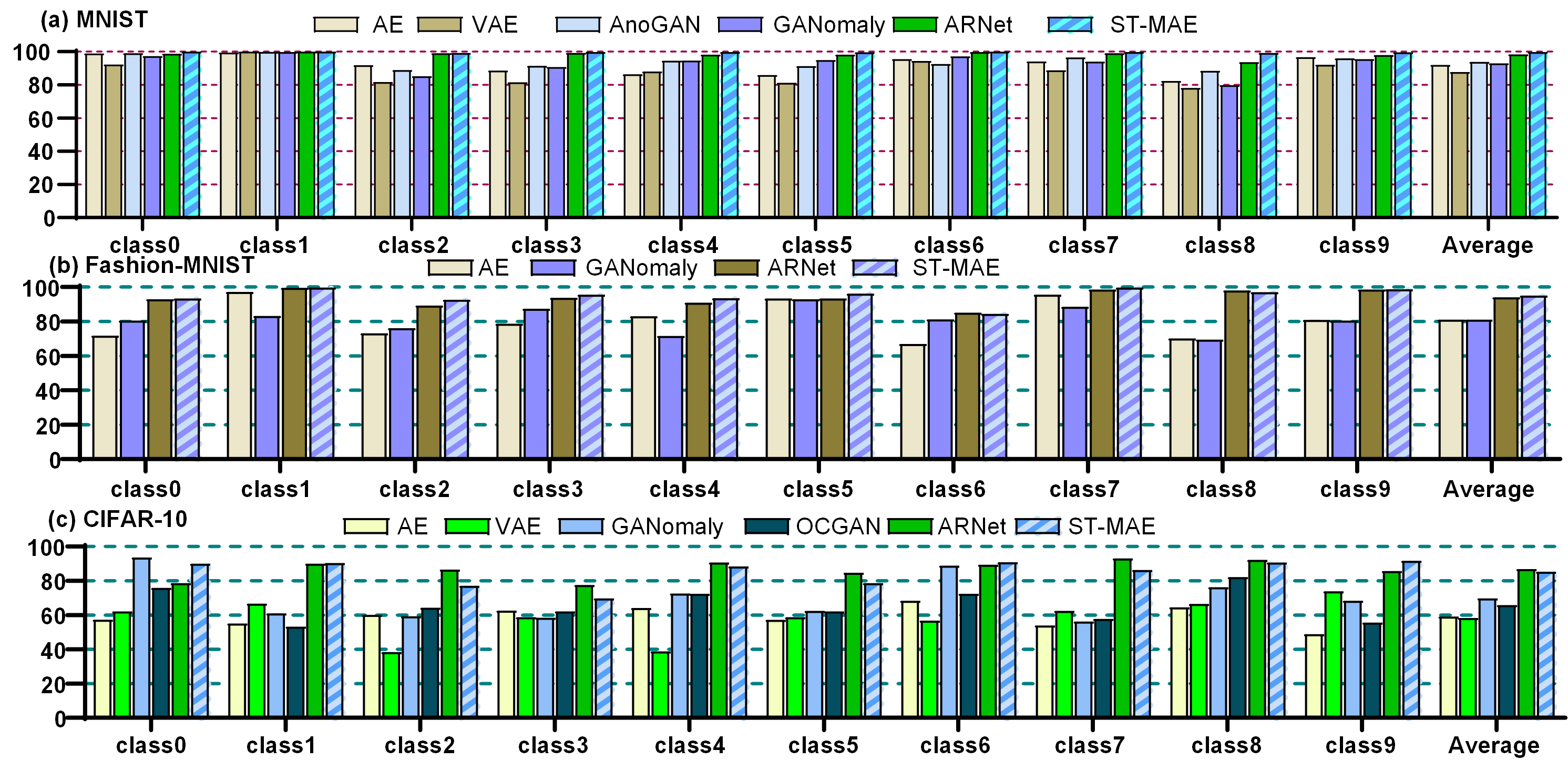}}
\caption[width=\textwidth]{Detection performance (AUC ROC) on the three one-class novelty detection datasets. The class numbers on the x-axis denote the class that is assumed to be normal. (a) MNIST dataset. (b) Fashion-MNIST dataset. (c) CIFAR-10 dataset.
}
\label{fig24}
\end{figure*}
Then, to further validate the effectiveness of our ST-MAE, we visualized the qualitative localization results comparison in Fig. 9. The lesions localization performance of various methods including MemAE\cite{r13}, AE\cite{r13}, GANomaly\cite{r57}, RIAD\cite{r7}, Patch-SVDD\cite{r9}, SPADE\cite{r4} and GCPF\cite{r5} on three types of lesions and the normal case are given. As shown, our ST-MAE can simultaneously detect different types of lesions and is capable of suppressing false detection noise in normal areas. Moreover, the image-level anomaly score histogram of the Retinal-OCT dataset by the proposed ST-MAE method is exhibited in Fig. 10, anomaly samples would get higher anomaly scores while the normal ones do the opposite. The distinction demonstrates that our method can accurately 
 discriminate the normal and abnormal samples utilizing the anomaly score.

% The P-net utilizes the structural information of retinal images that sourced from the pre-trained segmentation network supervised by the pixel-level annotations for training.

\subsubsection{Case study 3: Intelligent Video Surveillance} Experiments on the video anomaly detection tasks are conducted to assist in proving the generalizability of our approach. Most recent benchmarks for video anomaly detection exploit the motion information to understand the normal spatiotemporal patterns\cite{r47}. Since ST-MAE is not specialized for video anomaly detection, it relies on the individual frame and does not exploit motion information, therefore it does not perform as well as existing state-of-the-art methods. As shown in  Table V, we consider Conv2D-AE\cite{r49}, Conv3D-AE\cite{r50}, LSTM-AE\cite{r51} and ConvLSTM-AE\cite{r48} as the baselines, FFP\cite{r52} and SmithNet\cite{r53} as the SOTAs. Overall, our ST-MAE outperforms all the baseline models and achieves comparable results to some state-of-the-art methods, denoting the capacity for further improvement in the field of video anomaly detection by utilizing motion information. 

Some of the abnormal event localization maps within the frame are visualized on the right side of Fig. 11, the anomaly maps indicate that our ST-MAE has a remarkable ability for precise abnormal event localization. The left side of Fig. 11 shows the frame-level abnormal score curves of the proposed ST-MAE on two video samples, where the highlighted interval indicates the presence of abnormal events. The results illustrate the sensitive tracking capability of our ST-MAE for abnormal events.

% \begin{table*}
% \caption{{Image-level/pixel-level AUC ROC results of different methods in MVTec AD}}
% \label{table}
% \setlength{\tabcolsep}{3pt}
% \begin{threeparttable}
% \begin{tabular}{p{\textwidth}}
% $\includegraphics[width=\textwidth]{Fig23.png}$
% \end{tabular}
% \end{threeparttable}
% \label{table4}
% \end{table*}
\begin{figure}[t]\centering
\includegraphics[width=8.8cm]{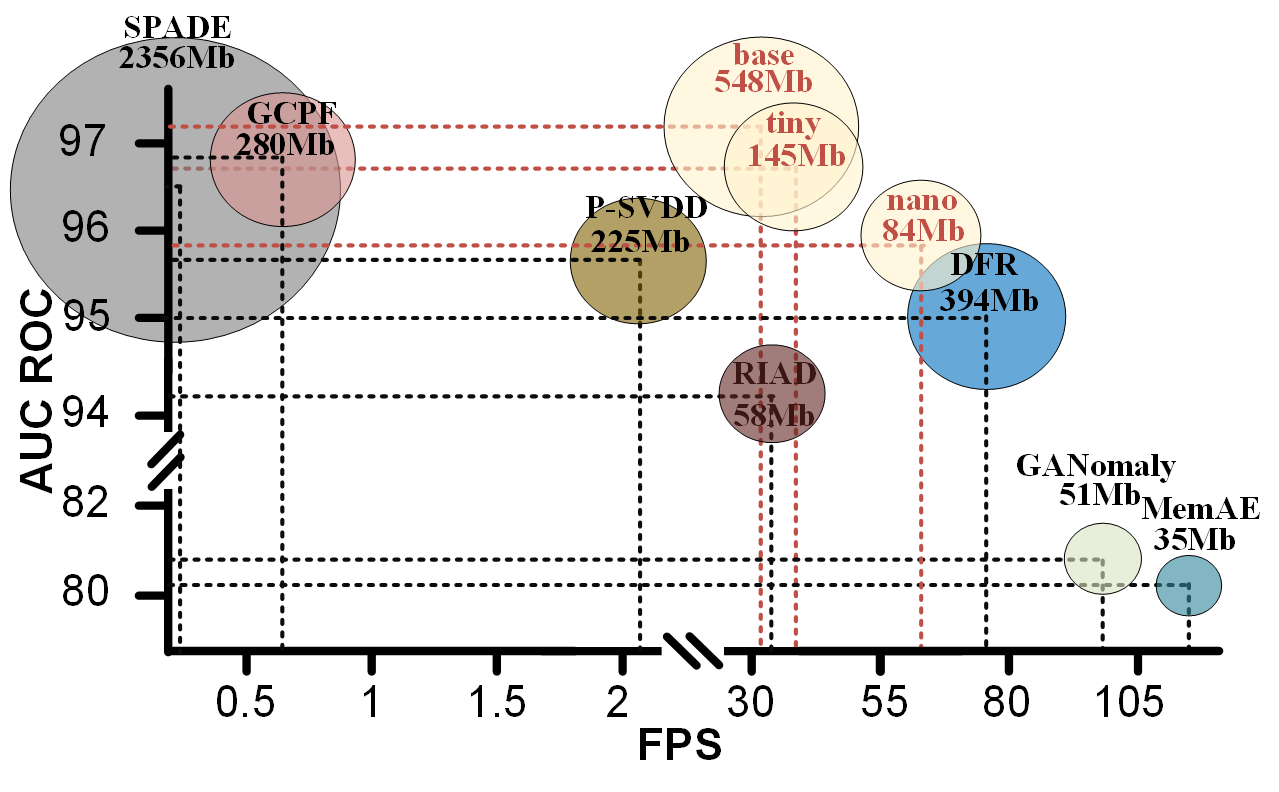}
\caption{Computational consumption comparison of various methods. The horizontal axis indicates the time consumption(FPS), the vertical axis represents the detection performance(AUC ROC), and the size of circulars denotes the GPU memory usage.}
\label{FIG_3}
\end{figure}

\subsubsection{Case study 4: Semantic one-class novelty detection} 

Although our ST-MAE method is mainly used for visual anomaly localization, experiments on the semantic one-class novelty detection task are also performed to prove the capability of ST-MAE to cope with the UVAD. In Fig. 12, the ST-MAE is involved in comparison with some state-of-the-art methods on the three intensively used benchmarks MNIST, Fashion-MNIST, and CIFAR-10. Our ST-MAE obtains competitive performances on the three datasets with the AUC ROC scores of 99.6, 94.9, and 85.1 respectively. substantially outperforming the widely used baseline AE\cite{r33}, VAE\cite{r63} and GANomaly\cite{r57}, and is comparable to optimal methods ARNet(98.3, 93.9, and 86.6). ARNet leverages the attributes restoration of the whole image to learn the overall normal semantics, which is specifically suitable for one-class novelty detection. However, its performance will be greatly reduced when dealing with abnormal localization. For example, it only obtained an AUC ROC of 83.9 on the Mvtec AD dataset, which indicates that ARNet is inadequate and effective in the real world. On the contrary, our ST-MAE not only advances state-of-the-art performance on real-world anomaly localization but also gets fairish results for the topic of semantic one-class novelty detection, denoting the potentiality to be the unified model for UVAD.

\subsection{Computational consumption comparsion}
As for the validation of the proposed ST-MAE inference effectiveness, We compare the computational consumption of the proposed framework, including frames per second (FPS) and GPU memory usage, with that of several state-of-the-art models. For a fair comparison, the computational consumption indicators for each method were measured for an image with a resolution of 256 $\times$ 256 and under the same hardware conditions. 

We studied the performance of three model variants of ST-MAE, and the comparison results are presented in Fig. 13. Among all the methods, the base model of ST-MAE reached the best AUC ROC performance for pixel-level localization. The time consumption of it is only 33.2 ms, which is only behind the DFR, GANomaly, and MemAE, while the performance markedly surpasses them in terms of detection accuracy. Furthermore, contrasted to the time-consuming discriminative embedding-based SOTAs, The ST-MAE(base) has a significant advantage in inference time. For example, with better AUC ROC performance, ST-MAE has almost 600, 30, and 15 times improvement over the SPADE, GCPF, and Patch-SVDD in terms of the computation time. As for the GPU memory consumption, the ST-MAE(base) takes 548 Mb of GPU memory, which is slightly more than other methods but still considerable compared with the SPADE (2356Mb). 

Moreover, for the lightweight model variants, the tiny and nano versions of ST-MAE can further improve operating efficiency and memory usage with a minor reduction in accuracy. Thus, ST-MAE can be easily adapted to hardware conditions with limited resources with distinguished performance, showing great potentiality in practical application.

\subsection{Ablation experiments}
\begin{table}
\caption{The AUC ROC of different image property representations on Mvtec AD dataset.}
\label{table}
\setlength{\tabcolsep}{3pt}
\begin{tabular}{p{\columnwidth}}
$\includegraphics[width=\columnwidth]{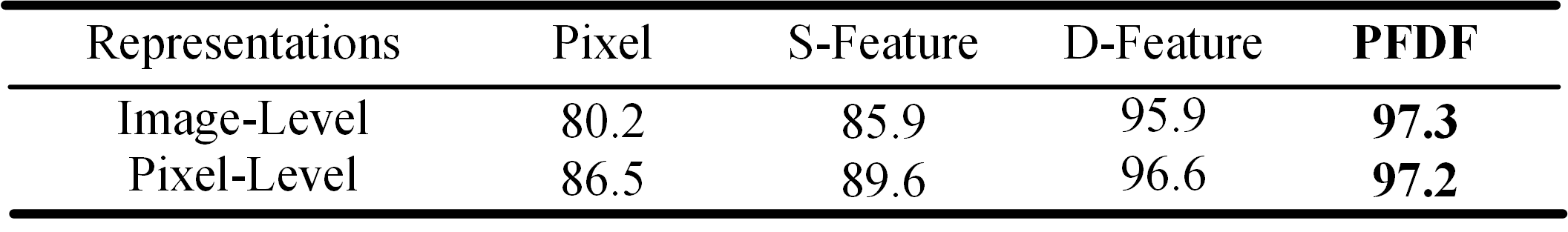}$
\end{tabular}
\label{table2}
\end{table} 

\subsubsection{Influence of the LPSR}

Our ST-MAE system follows a hybrid fashion integrated with DCNN for the following benefits: (1) As mentioned in \cite{r16}, the inductive bias of convolutional networks is helpful for smaller datasets. (2) The deep features are more discriminability than the low-level pixel intensity. (3) The feature reconstruction can be viewed as the relaxation version of image reconstruction which enables more robust anomaly detection. Thus, the pre-trained DCNN model is adopted to extract the semantic representation PFDF of the image. To illustrate the effectiveness of this design, we perform a validation experiment on the Mvtec AD dataset with different image property representations.

First, the influence of the hierarchical features is explored. A series of derived schemas were investigated including the shallow hierarchical feature representations(S-Feature) in the first two blocks of the VGG19 network, the deep hierarchical feature representations(D-Feature) of the third and fourth blocks, the low-level image pixels where the DCNN was totally discarded, and the proposed PFDF. The above four different image property representations were fed into the ST-MAE model to verify their performance. The results are shown in Table VI, it can be concluded that compared with deep features, the image pixel reveals the weakest discriminability. At the same time, as the hierarchy feature goes deeper, the performance of all categories improves in general, this phenomenon occurs because the deeper features yield higher discriminability and are equipped with more contextual semantics. Furthermore, as evident from Table VI, the multi-scale fusion strategy can improve performance. Therefore, the proposed ST-MAE model tends to be more robust and accurate with the deep-shallow features of joint fusion.

\begin{figure}[t]\centering
\includegraphics[width=8.8cm]{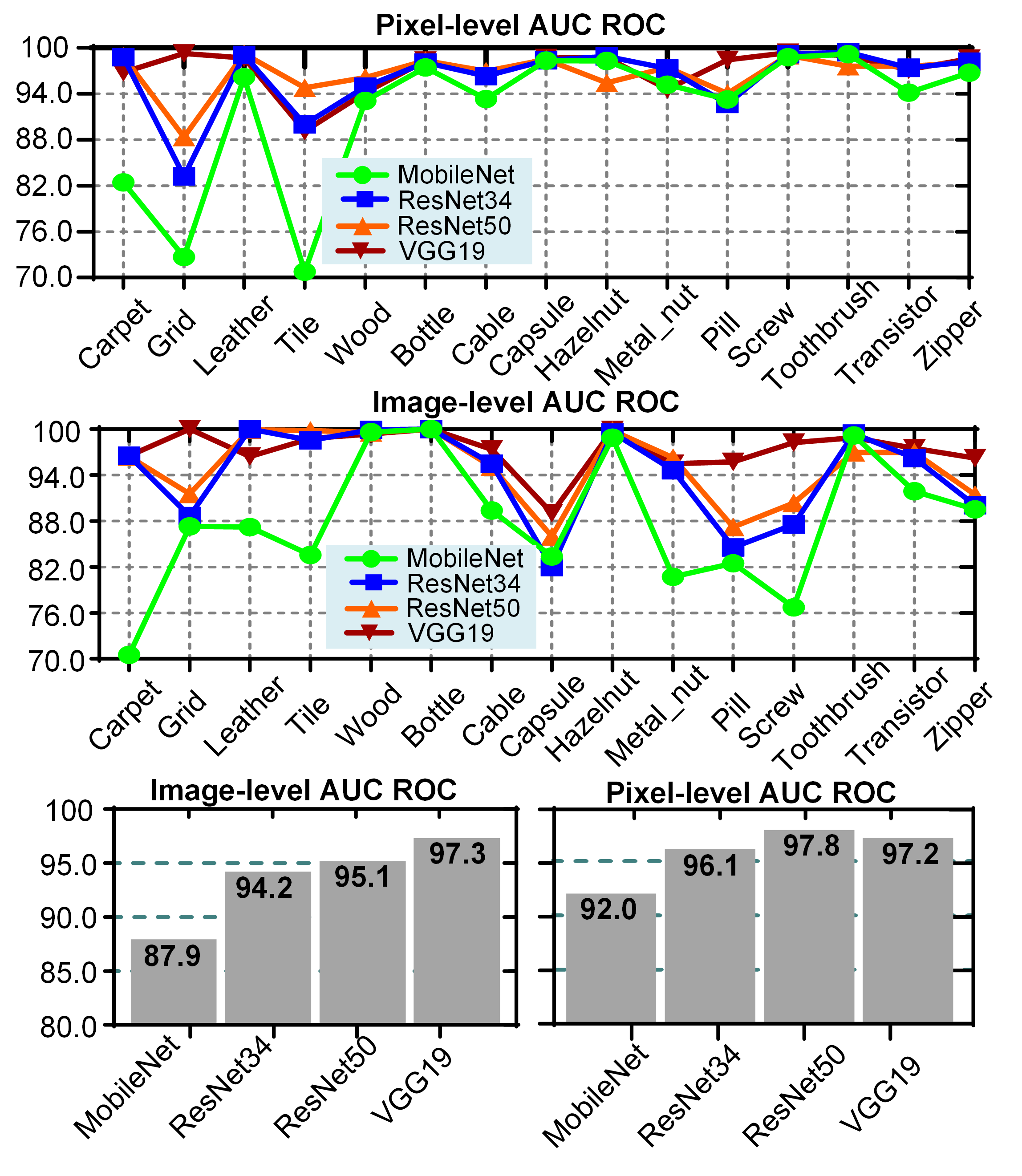}
\caption{The influence of the different pre-trained CNN models on image-level anomaly inspection and pixel-level anomaly localization performance.}
\label{FIG_3}
\end{figure}

Furthermore, the DCNN model types are also experimentally studied, including the Mobilenetv2, ResNet34, ResNet50, and the applied VGG19 \cite{r15}, and please note that the hierarchical features are extracted in the first four convolutional blocks for all models. As described in Fig. 14, the performances of the different DCNN models over 15 categories of the Mvtec AD dataset are displayed. It can be observed that all above backbones achieved preferable performance, reflecting that our ST-MAE can adapt to different kinds of backbones. Overall, the best result was obtained with ResNet50 and VGG19, while the lightweight Mobilenetv2's performance is slightly inferior. This result can be interpreted as different networks having different semantic representations for the feature embeddings, and the deeper networks have stronger representations. However, the volume of the extracted deep features with a too-deep network will exceed the reconstruction capacity of the ST-MAE model, which causes the ST-MAE to be unable to effectively reconstruct the normal features. Thus, the image level inspection performance of ST-MAE shows a slight degradation when cooperating with the ResNet50 model. Moreover, as for the selection criteria, it is still encouraged to obtain a good trade-off between representational power and computational complexity when applying proposed ST-MAE in practice.

\subsubsection{Impact of model scale}
As mentioned earlier, we designed the ST-MAE with four model variants of different scale, which can provide various options depending on hardware conditions and efficiency requirements.  Different model variants are experimentally studied in this section. The model performance and the total parameters(/Mb) are exhibited In Fig. 15, it can be observed that the performance improves as the scale of the model increases, while the huge model, on the other hand, has some performance degradation due to the fact that it requires more optimization time. In addition, the models with small parameters(Nano and Tiny) can also achieve acceptably favorable results. Thus, when it is involved in the actual deployment, the model variants should be adopted properly considering both efficiency and performance for different tasks, to fully utilize the performance of the ST-MAE property.

\begin{figure}[t]\centering
\includegraphics[width=8.8cm]{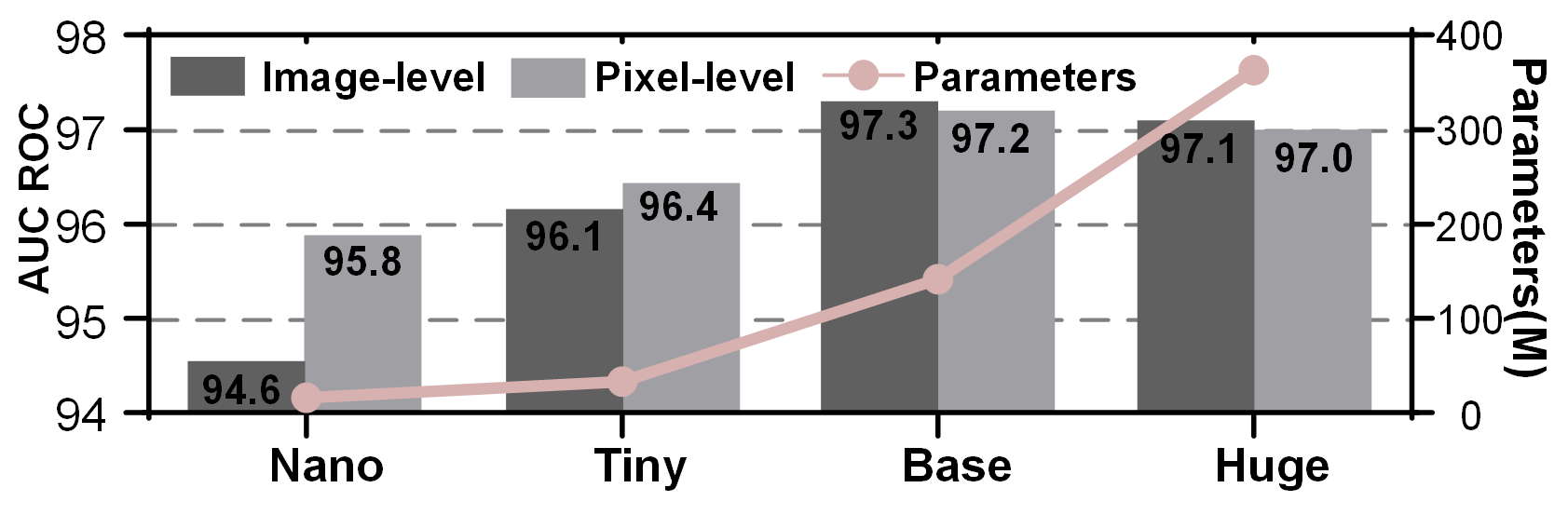}
\caption{The AUC ROC and the total parameters comparison results between the different model variants.}
\label{FIG_3}
\end{figure}

\begin{figure}[t]\centering
\includegraphics[width=8.8cm]{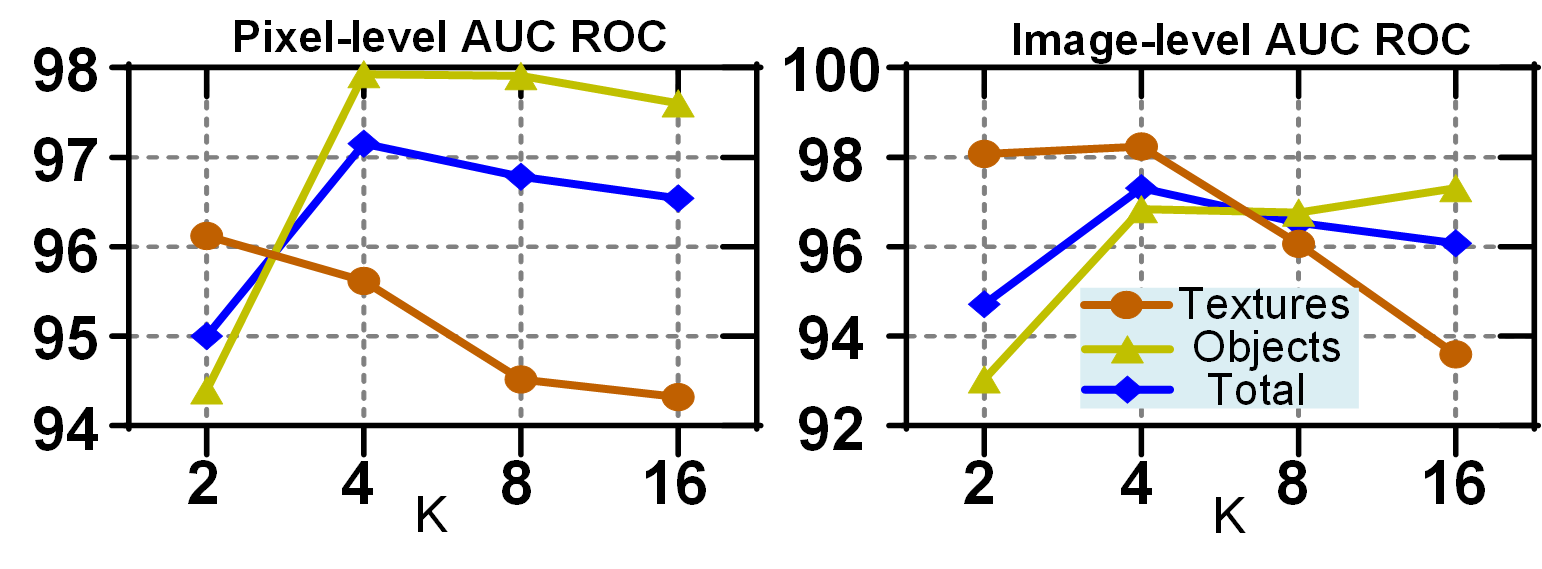}
\caption{The influence of the $K$ on pixel/image-level detection performance.}
\label{FIG_3}
\end{figure}

\begin{figure}[t]\centering
\includegraphics[width=8.8cm]{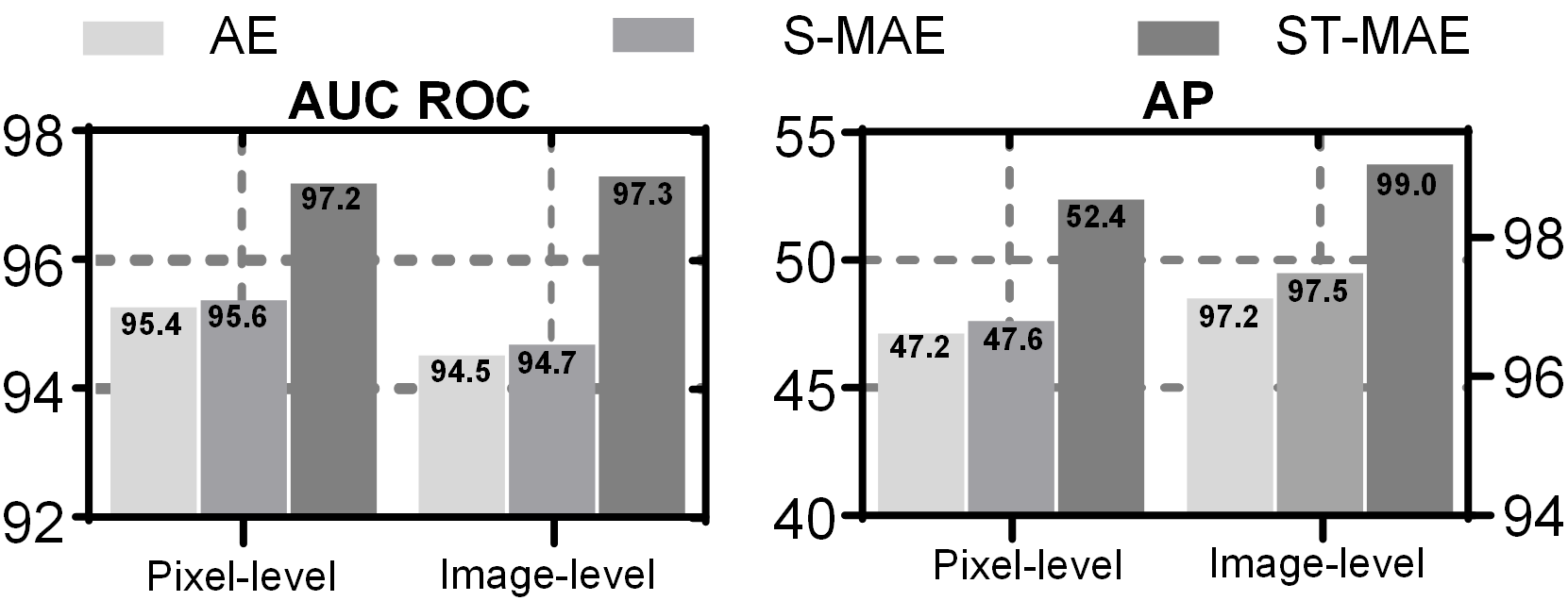}
\caption{The AUC ROC and AP comparison results between three ablation models on the MVtec AD dataset.}
\label{FIG_3}
\end{figure}

\subsubsection{Sensitivity of the patch size $K$ in FBTD}
Unlike the primary operation in the hybrid architecture of ViT\cite{r16} that uses the 1×1 as the spatial size of feature patches. In the FBTD, the 2D CNN feature map PFDF was split into a grid of patches with the patch size $K$. Each patch represents the local structural semantics of the input image, and the patch size $K$ controls the granularity of the dividing grid in the FBTD. Each feature patch token in the grid consists of a square of $K$ × $K$ feature vectors and is the basic unit in the sequence. If the patch size $K$ is smaller, the sequence length will increase accordingly, which makes the transition of two deep latent feature subsets more difficult; this setting will lead to inferior transition accuracy of normal latent representations. However, with a too-large pact size $K$, the partial information contained in each subset will become sparse, also being infeasible to obtain accurate transition results. Thus, the patch size $K$ will influence the transition procedure of two feature patch subsets and the further anomaly detection performance.
\begin{table}
\caption{Image/pixel-level AUC ROC comparison results for different loss modalities}
\label{table}
\setlength{\tabcolsep}{3pt}
\begin{tabular}{p{\columnwidth}}
$\includegraphics[width=\columnwidth]{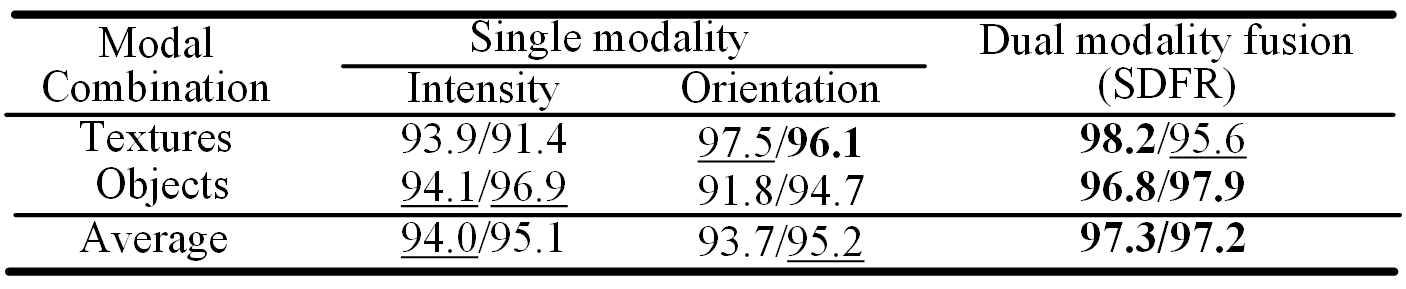}$
\end{tabular}
\label{table2}
\end{table}
To investigate its influence, we experiment with the effect of hyper-parameter $K$ in the FBTD module. As shown in Fig. 16, the image-level and pixel-level AUC ROC performances on the Mvtec AD dataset against the $K$ are exhibited. It can be seen that when the $K$ varies in the range $\{$2, 4, 8, 16$\}$, the image/pixel level AUC ROC first increases and then decreases as $K$ increases further. Furthermore, the inspection results of the texture category showed a monotonic decreasing trend with increasing $K$. This can be explained that the texture objects are more homogeneous thus ST-MAE is capable of handling long sequence transitions, its performance is only affected by the granularity of the patch division. For the object categories, the optimal performance is obtained at $K$=4. This result occurs because the structure of the object class varies significantly with its position, a smaller $K$ causes the sequence of two subsets to be longer, which makes the transition of two subsets difficult. However, with a too-large $K$ value, the transition accuracy of normal representations is reduced due to the lower granularity.

The overall performance is the best when $K=4$, moreover, it is worth noting that the smaller $K$ will lead to higher computational cost, which rises with a second order relative to the number of feature patch tokens. Thus we empirically set the $K=4$ as the optimal choice.

\subsubsection{Necessity of the Siamese transition}
The deep feature transition is introduced in the latent space of ST-MAE to cast the feature reconstruction as the feature transition, which alleviates the limitation of identity mapping of the trivial shortcut. It performs the mutual transition between two feature patch token sequences on the basis of the FPTD and the Siamese encoder. Two ablation models are designed and implemented to investigate how the Siamese encoder and the deep feature transition affect the final performance, namely the S-MAE where the deep feature transition is removed, and the AE where the Siamese encoder and the deep feature transition are all removed. In this situation, the whole system can be viewed as the basic Transformer based feature reconstruction model. The quantitative comparison of AUC ROC and Average Precision(AP) results of the above three ablation models are presented in Fig 17. When the Siamese encoder structure is introduced, the overall performance is slightly improved compared with the baseline AE with an improvement in AUC ROC of 0.2$\%$/0.2$\%$ and AP of 0.4$\%$/0.3$\%$. A greater improvement of a margin of 1.6$\%$/2.6$\%$ AUC ROC and 4.8$\%$/2.5$\%$ AP has been observed between the S-MAE and ST-MAE. The experimental results demonstrated that it is mainly the deep feature transition that alleviates the identity mapping shortcut, which will be discussed later. 
 
Thus, the Siamese transition mechanism can enhance the semantic extraction capability of the model by means of self-supervised learning, which enforces the model to learn the interior normal semantics patterns.

\subsubsection{Gains from the SDFR}

As stated before, the SDFR is introduced to measure the feature-level residual of intensity and orientational two-modal divergences to leverage robust anomaly detection. To illustrate its effectiveness, we experimented with the proposed ST-MAE to validate various anomaly score criteria. The evaluation results are listed in Table VII, It can be observed that the SDFR combining two loss functions outperforms the single-mode constraints in both texture and object categories. This phenomenon can be explained for the reasons: (1) Different anomaly patterns present different activation responses in different modalities, implying that the potential anomalies may be concealed in the unimodal method. (2) The accuracy of the unimodal methods may be limited by noise corruption, which will lead to false alarms in the background region. 

Through contrastive analysis of the SDFR and the other single-modal loss functions, it can be concluded that the SDFR with dual-modal residual fusion mechanism is able to promote detection robustness and accuracy.

\section{Extended experiments and analysis}

\subsection{The feature reconstruction results}
\begin{figure}[t]\centering
\includegraphics[width=8.8cm]{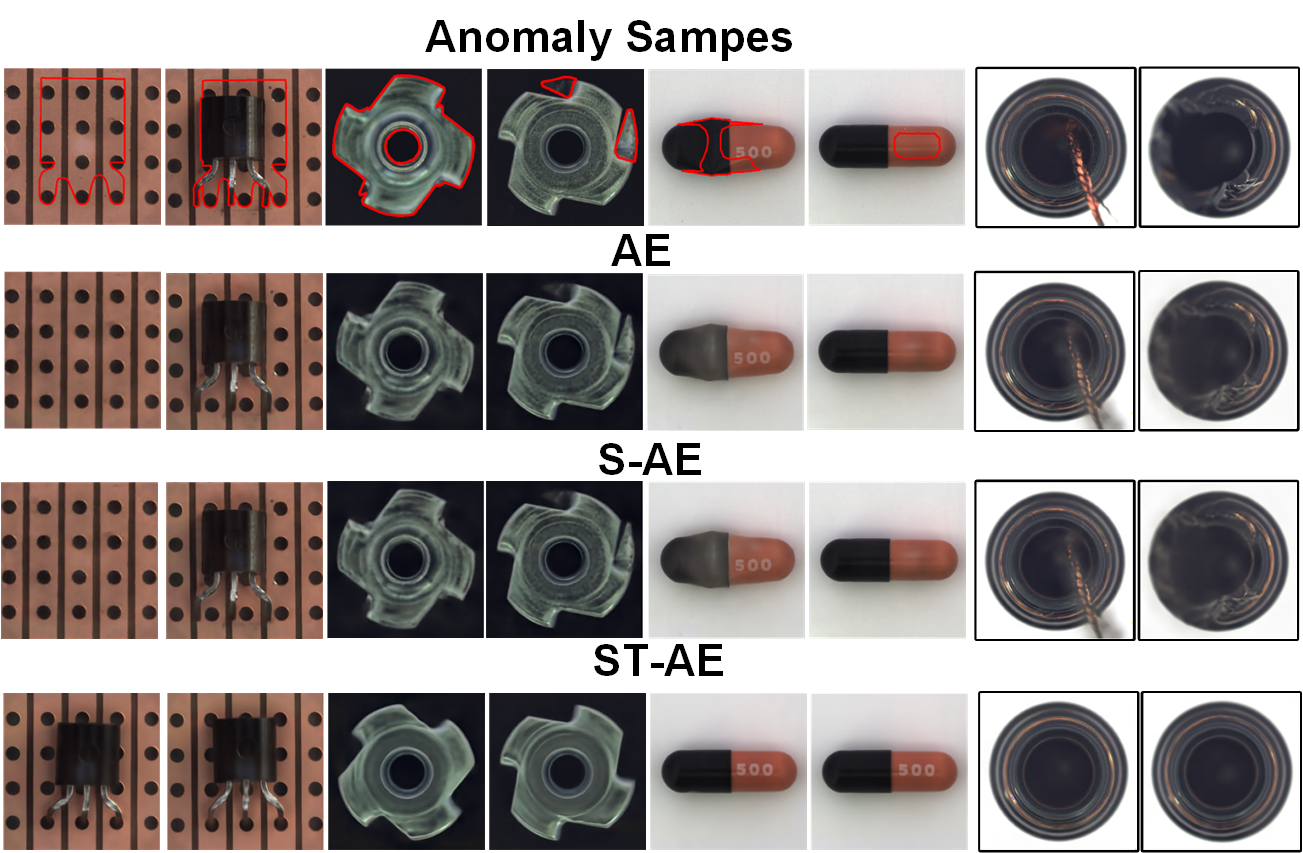}
\caption{Visualization results of reconstructed features of three different model variants.}
\label{FIG_3}
\end{figure}

The ST-MAE has achieved better performance by introducing the deep feature transition as the implicit self-supervisory tasks into anomaly detection. Here we further investigate how ST-MAE avoids the identity mapping of the trivial shortcut and extracts the semantic features by visualizing the reconstruction results. 
For reason that our ST-MAE performs a feature reconstruction schema, which is inconvenient to observe compared to image reconstruction. Thus, we introduce an auxiliary decoder to rend the reconstructed features of PFDF to the images. Specifically, this decoder uses a plain convolutional neural network design, and the overall structure is symmetric to VGG19. Notably, the auxiliary decoder is only used for visualization and has no effect on ST-MAE performance.

As can be seen in Fig. 18, the three ablation models mentioned in Section IV.E.(4) were investigated for their reconstruction behavior using the auxiliary decoder.  As we expected, S-MAE using a Siamese encoder structure does not address the drawbacks of identity mapping. However, ST-MAE with the further introduction of deep feature transition can solve this problem. As illustrated in Fig. 18, different types of anomalies can be reconstructed by ST-MAE as normal patterns. Regardless of the overall semantic deviation or the degradation of local regions. This property can effectively improve the performance of subsequent feature reconstruction residual-based anomaly detection.

\subsection{Extension to few-shot situations}

\begin{figure}[t]\centering
\includegraphics[width=8.8cm]{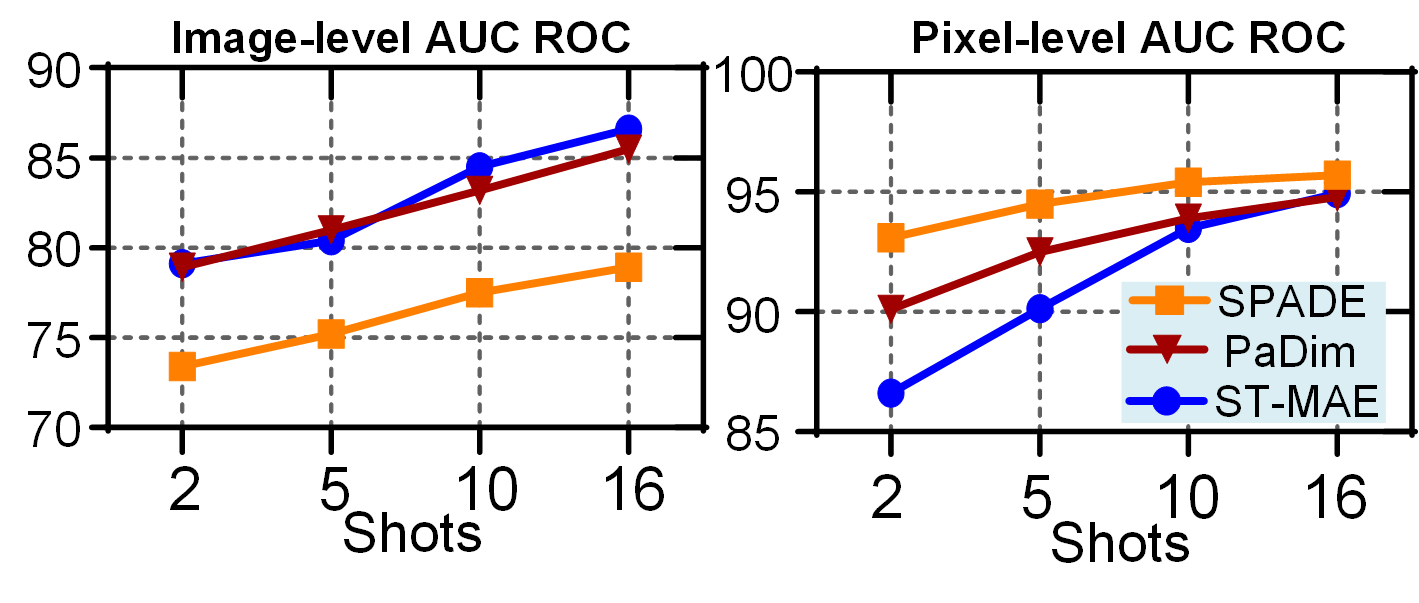}
\caption{The few-shot anomaly detection results of image-level and pixel-level AUC ROC for ST-MAE and competing methods on the MVtec AD dataset}
\label{FIG_3}
\end{figure}

As an extension of the experiments, we consider a practical but not widely studied scenario: few-shot anomaly detection where limited samples are available for training. Our model can be used in such scenarios due to the following properties:(1) Instead of using the pixel reconstruction, we adopted the relaxed version: deep feature  reconstruction which can significantly improve robustness and reduce model requirements in the limited sample case. (2) The proposed deep feature transition can construct multiple tasks based on a limited sample. In the deep feature transition strategy, the full set of deep feature patch tokens are randomly decoupled into two subsets, which are subsequently transitioned to each other. We here define the mutual transition of a pair of complementary subsets as a task, such that a sample can yield a large number of transition tasks($C_{N^2}^{\frac{N^2}{2}}$), which can significantly improve the robustness of the model with few samples. In this subsection, we experimented our ST-MAE on the few-shot setting on the Mvtec AD dataset to compare with existing advanced few shot anomaly detection approaches SPADE\cite{r4} and PaDim\cite{r11}. The experimental results are shown in Fig. 19. Under the few shot setting,  ST-MAE achieves 79.1 $\%$, 80.4$\%$, 84.5$\%$, and 86.6$\%$ in average image-level AUC ROC and 86.6.1 $\%$, 90.1$\%$, 93.5$\%$, and 94.9$\%$ in average pixel-level AUC ROC on MVTec AD dataset with 2-shot, 5-shot, 10-shot, and 16-shot scenarios, respectively.

Compared with SPADE which uses the normal feature retrieval and PaDim which employs the normal feature distribution, our ST-MAE method performs comparably to the above two at the image level AUC ROC and less well at the pixel level AUC ROC, our ST-MAE is more efficient by directly utilizing the residuals of feature reconstruction. 

\section{Conclusion}
In this paper, we presented the ST-MAE, a DCNN-Transformer hybrid model for Unified Visual Anomaly Detection(UVAD). The ST-MAE first leverages the pre-trained DCNN to extract the semantic representation of the input image by LPSR. Then, the FBTD is utilized to construct two complementary feature patch token sequences. Subsequently, the ST-MAE is a novelty proposed to perform the mutual transition of two sequences. Finally, the anomaly score map can be obtained by the SDFR.
We experimented with ST-MAE on multiple benchmarks, which cover almost all types of application scenarios currently available. The comparative experiment results reveal its state-of-the-art performance, excellent generalization adaptability, and the hardware resource-friendly efficient inference process. The ablation analysis experimental results effectively indicate the contribution of each module of the proposed method. Finally, we investigated its reconstruction effect in solving the identity mapping of the trivial shortcut and further explore the performance in few-shot scenarios. In the future, We will validate this approach in more fields and further improve it.
% References
\bibliographystyle{ieeetr} 
\bibliography{reference}

\begin{IEEEbiography}[{\includegraphics[width=1in,height=1.25in,clip,keepaspectratio]{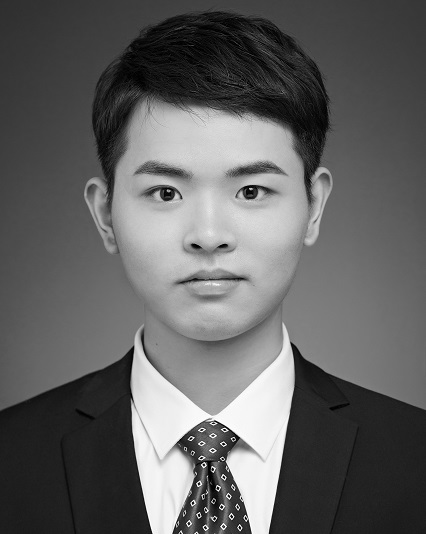}}]{Haiming Yao} received a B.S. degree from the School of Mechanical Science and Engineering, Huazhong University of Science and Technology, Wuhan, China, in 2022.
He is pursuing a Ph.D. degree with
the Department of Precision Instrument, Tsinghua
University.

His research interests include visual anomaly detection, deep learning and  edge intelligence.
\end{IEEEbiography}

\begin{IEEEbiography}[{\includegraphics[width=1in,height=1.25in,clip,keepaspectratio]{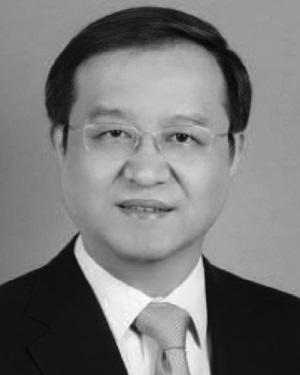}}]{Xue Wang}received an M.S. degree in measurement and instrumentation from Harbin Institute of Technology,
Harbin, China, in 1991 and a Ph.D. degree in mechanical engineering from Huazhong University of Science and Technology, Wuhan, China, in 1994.

He was a Postdoctoral Fellow in electrical power systems with Huazhong University of Science and Technology from 1994 to 1996. He then joined
the Department of Precision Instrument, Tsinghua University, Beijing, China, where he is currently a Professor. From May 2001 to July 2002, he was a Visiting Professor with the Department of Mechanical Engineering, University of Wisconsin--Madison. His research interests include topics in wireless sensor networks, cyber-physical systems, intelligent biosignal processing, medical image processing, and smart energy utilization.

Prof. Wang is a senior member of the IEEE Instrumentation and Measurement Society, Computer Society, Computational Intelligence Society, and Communications Society.
\end{IEEEbiography}

\begin{IEEEbiography}[{\includegraphics[width=1in,height=1.25in,clip,keepaspectratio]{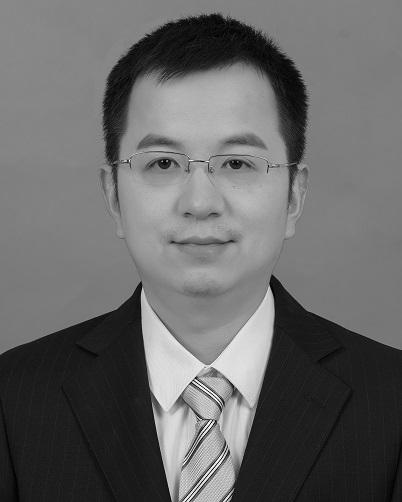}}]{Wenyong Yu} received an M.S. degree and a Ph.D. degree from Huazhong University of Science and Technology, Wuhan, China, in 1999 and 2004, respectively. He is currently an Associate Professor with the School of Mechanical Science and Engineering, Huazhong University of Science and Technology. 

His research interests include machine vision, intelligent control, and image processing.
\end{IEEEbiography}

\end{document}